%% file: main.tex
\documentclass[10pt,twocolumn,letterpaper]{article}

\usepackage{cvpr}              

\input{preamble}

\usepackage{xcolor}
\definecolor{plotblue}{HTML}{1f77b4}
\definecolor{plotorange}{HTML}{ff7f0e}
\definecolor{plotgreen}{HTML}{2ca02c}

\usepackage{algorithm}
\usepackage{algpseudocode}
\usepackage{booktabs}
\usepackage{array} 
\usepackage{ragged2e} 
\usepackage{makecell} 
\usepackage{multirow}
\usepackage{float}

\definecolor{cvprblue}{rgb}{0.21,0.49,0.74}
\usepackage[pagebackref,breaklinks,colorlinks,allcolors=cvprblue]{hyperref}

\def\paperID{36292} 
\def\confName{CVPR}
\def\confYear{2026}

\title{Grounding Hierarchical Vision-Language-Action Models Through Explicit Language-Action Alignment
}

\author{Theodor Wulff\thanks{\small Corresponding author: theodor.wulff@manchester.ac.uk\\
© 2026 IEEE.  Personal use of this material is permitted.  Permission from IEEE must be obtained for all other uses, in any current or future media, including reprinting/republishing this material for advertising or promotional purposes, creating new collective works, for resale or redistribution to servers or lists, or reuse of any copyrighted component of this work in other works.}
\and Federico Tavella  
\and Rahul Singh Maharjan 
\and Manith Adikari 
\and Angelo Cangelosi \\ 
\\
The University of Manchester\\
Manchester, United Kingdom\\
}

\begin{document}
\maketitle
\input{sec/0_abstract}
\input{sec/1_intro}
\input{sec/2_relatedWork}

\input{sec/3_problem}
\input{sec/4_method}
\input{sec/5_experiments}

\input{sec/6_results}
\input{sec/7_discussion}
\input{sec/8_conclusion}
{
    \small
    \bibliographystyle{ieeenat_fullname}
    \bibliography{main}
}
\clearpage
\input{sec/9_appendix}

\end{document}

%% file: preamble.tex
\renewcommand{\paragraph}[1]{\vspace{.5em}\noindent\textbf{#1.}}



%% file: sec/0_abstract.tex
\begin{abstract}
Achieving robot transparency is a critical step toward effective human-robot collaboration.
To be transparent, a robot's natural language communication must be consistent with its actions and explicitly grounded in the task and environment.
Existing hierarchical Vision-Language-Action (VLA) models can generate language (e.g., through chain-of-thought) and low-level actions.
However, current work does not consider explicit alignment between these modalities during training.
To address this crucial gap, we propose a novel training framework that explicitly grounds hierarchical VLA sub-task descriptions with respect to the visual observation and action space. 
Our framework uses a contrastive model to assess the alignment between generated language and corresponding action trajectories.
This contrastive model enables direct ranking of different language-trajectory pairs based on their alignment, allowing us to refine the grounding of our hierarchical VLA through offline preference learning.
We apply our framework to the LanguageTable dataset, a benchmark dataset of human language-annotated trajectories, and provide critical insights into multimodal grounding representations, all while establishing a strong baseline that achieves performance comparable to fully supervised fine-tuning and minimizing the need for costly data annotations.
\end{abstract}

%% file: sec/1_intro.tex
\section{Introduction}
Humans use natural language to express thoughts, observations, and actions, enabling us to collaborate across unseen scenarios and environments.
During human-robot collaboration, humans expect robot behavior to be explained, like how humans provide explanations to other humans, offering brief explanations on why it failed and its behavior~\cite{han2021VErbal}. 
Thus, embodied AI systems need to be equipped with expressive communication abilities to pave the way for effective human-robot collaboration. 
The central challenge lies in ensuring that a robotic agent’s communication is not only coherent in language, but also faithfully grounded in its visual perception and behavior.
Specifically, this requires solving the problem of symbol grounding~\cite{harnadSymbolGroundingProblem1990}. 
In the context of robotics, grounding describes the ability to relate textual descriptions and visual features to spatial object relations and actions in the real world~\cite{ichterCanNotSay2023}.

Robotic research is one of many disciplines that is being boosted by the rise of Large Language models (LLM) and Vision-Language models (VLM), affecting robotic tasks like plan generation or control policies for robot actions~\cite{ma2024survey}.
Current work leverages the effective processing capabilities of VLMs used in Vision-Language-Action models (VLA) for generative tasks that produce diverse outputs such as text, images, or continuous actions. 
VLAs benefit from the generalizability of the VLM across a diverse distribution of language and image inputs.
In most cases, their task is to produce actions either by generating discrete tokens as if they represented word tokens~\cite{kim2025openvla} or by utilizing specific decoder heads for continuous action generation~\cite{black2024p0VisionLanguageActionFlow,gr00tn1_2025,palo2024keypoint}.
However, VLAs leverage powerful VLM backbones, providing an untapped capacity for rich, high-quality language output. 

Prompting LLMs and VLMs to think step-by-step, commonly referred to as chain-of-thought prompting~\cite{weiChainofThoughtPromptingElicits2022}, has led to improved results in commonsense, arithmetic and symbolic reasoning capabilities.
Alongside performance gains, this technique also benefits the human prompting the model, who often can follow the reasoning leading to an answer more easily~\cite{weiChainofThoughtPromptingElicits2022,zhou2023leasttomost}.
Similarly, hierarchical VLAs~\cite{belkhaleRTHActionHierarchies2024,shiHiRobotOpenEnded2025,blackpi05VisionLanguageActionModel2025} take a high-level task as input and produce an executable subtask as intermediate output, much like chain-of-thought prompting in LLMs.
This intermediate sub-task output provides model transparency, by semantic grounding and expression in natural language.
However, both the generated sub-steps and the final result carry no certainty of being error-free~\cite{NEURIPS2023_ed3fea90}.
Consequently, this introduces an additional layer of generation noise in hierarchical models, since the final output now depends on self-generated, decoupled intermediate outputs~\cite{shiHiRobotOpenEnded2025}.
This intermediate sub-task output provides model transparency, by semantic grounding and expression in natural language; at the cost of the additional problem of learning to align the language and action modalities~\cite{salehzade2022purposefulComm}. 

Learning objectives of current state-of-the-art VLAs are mainly concerned with the success rate of the model in a given task or the ability to generalize well to a large quantity of expert trajectories across embodiments. 
However, pure success rates fail to reveal how grounded the intermediate outputs are with respect to the actions and visual observation. 
One can only assume that the agent understood the instruction, but an evaluation metric that discloses possible misunderstanding is missing.
To address this, we propose to train an action-conditioned grounding model on human-annotated language-action-vision data through contrastive learning to explicitly evaluate the grounding of language with respect to actions and visual observations. 
We test this approach in a benchmark task for VLAs, where a robot, provided with a high-level task description, must generate both sub-task descriptions and physical pushing actions to arrange objects.

Inspired by recent work that uses learned rewards for preference optimization~\cite{zhang2024grapegeneralizingrobotpolicy}, we propose a novel framework, called GPLA, for \textbf{G}rounded \textbf{P}reference-based \textbf{L}anguage-action \textbf{A}lignment, to improve the grounding and transparency of hierarchical VLAs. 
To circumvent the expensive annotation of intermediate outputs, our approach first trains a contrastive model to generate an explicit grounding score based on the alignment between generated natural language sub-goals, visual observations, and subsequent actions.
We then use this learned score to generate preference pairs (i.e., more grounded vs. less grounded outputs) and employ preference-based fine-tuning to align the VLA towards producing more semantically correct intermediate steps.
We demonstrate our framework on the LanguageTable manipulation benchmark~\cite{lynchInteractiveLanguageTalking2022}, providing critical insights into the quality and limitations of language grounding representations within hierarchical VLAs.

To summarize, our contributions are the following:
\begin{itemize}
    \item We propose GPLA, a novel framework that uses preference learning to directly ground the intermediate language outputs of hierarchical VLAs with visual observations and actions, potentially eliminating the need for expensive annotation collection of intermediate outputs.
    \item GPLA achieves comparable performance with respect to the generated trajectories on the LanguageTable manipulation benchmark~\cite{lynchInteractiveLanguageTalking2022} to fully supervised fine-tuning, while uniquely offering applicability to low-data regimes. 
    \item Visual analysis of the embedding space shows that pre-trained self-supervised models can be used to generate an explicit grounding score. 
    However, they show clear distinctions between the visual observations and language inputs; our action-conditioned grounding model improves this separation by mapping action-vision and text inputs into overlapping embedding space.
\end{itemize}

%% file: sec/2_relatedWork.tex
\begin{figure*}[ht]
    \centering
    \includegraphics[width=\linewidth]{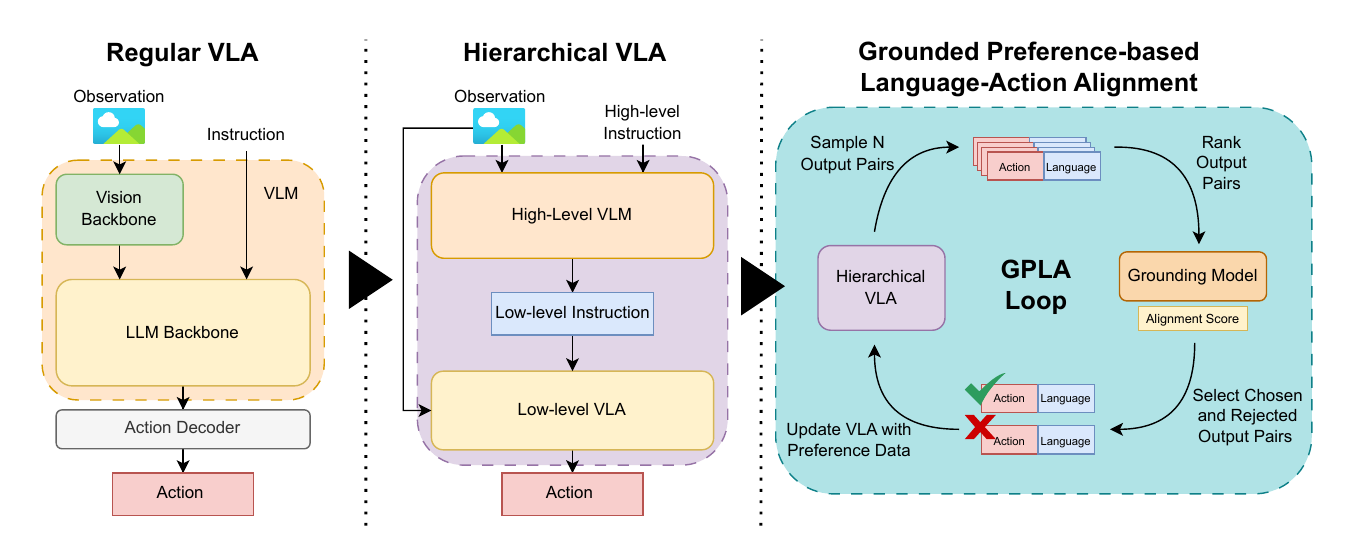}
    \caption{\textbf{Method Overview.} 
    We extend a regular VLA (left) into a hierarchical VLA by adding a high-level VLM module to break a high-level instruction down into executable low-level instructions (center), following recent trends on hierarchical VLAs~\cite{shiHiRobotOpenEnded2025,belkhaleRTHActionHierarchies2024}.
    To align the intermediate low-level instruction and the generated trajectory, we invoke a separately trained ranking model, which ranks $N$ sampled output pairs based on their grounding in the environment. 
    Based on these scores, we select the chosen and rejected output pairs that serve as preference data to update our transparent VLA and increase the alignment of the multimodal outputs.
    }
    \label{fig:method-overview}
\end{figure*}
\section{Related Work}
\subsection{Vision-Language-Action Models}
Foundation models are trained on large-scale data and have achieved outstanding performances across tasks in natural language processing, computer vision, and other domains~\cite{awaisFoundationModelsDefining2025}. 
The same paradigms have started to achieve generalizable results in the field of robotics, where huge collections of data~\cite{oneill2024openx} enabled the training of foundation models for action prediction~\cite{brohanRT1RoboticsTransformer2023b,brohanRT2VisionLanguageActionModels2023a,kim2025openvla,black2024p0VisionLanguageActionFlow,blackpi05VisionLanguageActionModel2025,belkhaleRTHActionHierarchies2024,octomodelteamOctoOpenSourceGeneralist2024,gr00tn1_2025}.
These models are known as Vision-Language-Action models (VLA) and generate robot actions given a visual observation and a language instruction.
Promising approaches to use the Vision-Language Model backbone to generate actions vary between discrete action token generation~\cite{kim2025openvla,brohanRT2VisionLanguageActionModels2023a} or using a specific decoder head to produce continuous action vectors using a learned linear mapping or a diffusion head~\cite{brohanRT1RoboticsTransformer2023b,black2024p0VisionLanguageActionFlow,blackpi05VisionLanguageActionModel2025,belkhaleRTHActionHierarchies2024,octomodelteamOctoOpenSourceGeneralist2024,gr00tn1_2025}. 
Most of these models utilize a Large Language model to process the visual and text-based inputs. 
However, many VLAs make little use of the full vocabulary of these models, since the relevant output only exists in the action domain.
Other works use hierarchical architecture to produce language to verbally respond using a VLM and create a plan, which is then passed to another VLA~\cite{shiHiRobotOpenEnded2025} or dedicated decoder~\cite{belkhaleRTHActionHierarchies2024} for policy execution. 

We identify two shortcomings in the current literature: first, these works rely on separately trained modules without a dedicated mechanism to further align the respective outputs; second, evaluation does not consider the quality of the intermediate outputs.

\subsection{Contrastive Representation Learning}
As a form of self-supervised representation learning, contrastive learning~\cite{balestriero2023cookbookselfsupervisedlearning} aims to extract information-rich embeddings from various input domains.
This is achieved by mapping inputs into a latent embedding space such that the distance between similar (positive) pairs is minimized and the distance between dissimilar (negative) pairs is maximized.
Contrastive learning has been effective in extracting general representations for various downstream tasks like classification or object detection~\cite{tschannenSigLIP2Multilingual2025,sontakke2023roboclip,radfordLearningTransferableVisual2021,heMomentumContrastUnsupervised,zhaiSigmoidLossLanguage2023,chenSimpleFrameworkContrastive2020}.

In Robotics, contrastive representation learning is leveraged to learn generalizable reward functions by establishing a similarity metric: approaches learn representations where latent similarities increase as an agent progresses towards the goal state.
The learned similarity scores between observations can then be used to reward an agent in a reinforcement learning scenario~\cite{ma2023liv,ma2025contrastive,sontakke2023roboclip,ma2023vipuniversalvisualreward,nair2022r3m,eysenbachContrastiveLearningGoalConditioned2022,bizaOnRobotReinforcementLearning2025,shao2020concept}.
The strength in contrastive representation learning lies in the ability to learn semantic similarities between modalities as long as positive and negative pairs are available. 
Besides rewarding task progression, such a reward model can also be used for specific evaluation of aspects like groundedness of the output~\cite{giannone2025supervisionfreevisionlanguagealignment}. 
To the best of our knowledge, there has not been an application of incorporating a grounding score into the training of hierarchical VLAs.

\subsection{Preference Learning} 
Learning from preferences is a key component in reinforcement learning from human feedback (RLHF), a method effectively applied to train modern Large Language Models~\cite{ouyang2022TrainingLanguageModels,openaiGPT4TechnicalReport2023}.
The original RLHF method~\cite{christiano2017rlhf} involves the explicit training of a reward model that learns the latent rewards of human preference distributions given a human-annotated dataset of chosen/rejected model outputs. 
Further efforts were made to remove the explicitly trained reward model and train the generative model directly on preference data~\cite{rafailov2023dpo,meng2024simpo,meta2024llama}.
Applications in robotics use preference learning to learn from teacher models or human preferences~\cite{lee2021pebble,myers2023activerewardlearningonline,iii2022fewshot}, rank trajectories based on different costs to extract preferences from a set of trajectories~\cite{zhang2024grapegeneralizingrobotpolicy,wang2024rlvlmf}, or by learning a reward function by ranking video frames based on their time step~\cite{yangRank2RewardLearningShaped2024}.

Preference learning has been shown to be effective in improving generative output in settings where quantifying model outputs is difficult~\cite{christiano2017rlhf,stiennonLearningSummarizeHuman2020}. 
We propose learning from ranked trajectories alongside transparent statements with respect to their quality, grounding in the environment, and language action alignment.

%% file: sec/3_problem.tex
\begin{algorithm*}
\caption{GPLA Grounded Preference-based Language-Action Alignment}\label{alg:training}
\begin{algorithmic}[1]
\Require Hierarchical VLA $\pi_\theta$ with high-level VLM $\pi_{\text{VLM}}$, dataset $D = \{ (x_i, o_i) \}$, grounding model $\pi_g$, batch size $B$, sampling limit $N_S$, iteration limit $N_I$
\Ensure Hierarchical VLA $\pi^*$ with improved action-language grounding
\State i = 0
\While{$i \le N_I$}
    \State Sample $B$ instruction-observation pairs $(x, o)$ from $D$ 
    \For{Each sample $(x, o)$}
        \State Generate $N_S$ language-action candidates $\{y_j\}_{j=1}^N$ using $\pi_\theta(x, o)$
        \State Compute grounding scores $g_j = \pi_g(x, o, y_j)$ for each $y_j$
        \State Select $y_{c} = \arg\max_j g_j$ and $y_{r} = \arg\min_j g_j$ as chosen/rejected pair
    \EndFor
    \State Update $\pi_{\text{VLM}}$ using SimPO loss on batch of $(y_c, y_r)$
    \State i += 1
\EndWhile
\end{algorithmic}
\end{algorithm*}
\section{Problem Formulation}
We define the problem of combined language action generation in line with previous work~\cite{wulff2025jointactionlanguagemodelling} by extending language conditioned behavior cloning (LCBC)~\cite{stepputtisLanguageConditionedImitationLearning2020a,nairLearningLanguageConditionedRobot2022a}.
LCBC can be defined as an imitation learning approach where an agent is conditioned to predict the subsequent action given an observation of the environment and task description in natural language~\cite{stepputtisLanguageConditionedImitationLearning2020a}.
The output is learned in a supervised manner, with expert action labels serving as ground truth.

We extend this problem by considering additional language output and putting constraints on the high-level task description. 
The high-level task description must be decomposable into multiple natural language sub-tasks. 
The overarching goal is to train a model that produces: 1) The subsequent action (the standard LCBC output) and 2) the current sub-task description a short-term, natural language summary of the current action).

Current hierarchical VLAs already provide the outputs required by this definition, since they often use a backbone VLM to break a high-level task down into an executable sub-task before generating the next action~\cite{belkhaleRTHActionHierarchies2024,shiHiRobotOpenEnded2025}. 
Current VLA action and language outputs are usually learned separately and lack explicit alignment; our framework establishes this explicit, grounded connection during training.

%% file: sec/4_method.tex
\section{Methodology}
GPLA requires a hierarchical VLA  ($\pi_\theta$) and a grounding model ($\pi_g$).
We construct the hierarchical VLA by integrating a pre-trained Gemma3 VLM, which decomposes high-level instructions, with a pre-trained VLA that generates executable action trajectories from the resulting low-level instructions (see Section~\ref{subsec:method_vla}).
Separately, we prepare a grounding model, which learns a shared embedding space across the vision, action, and language modalities, to assess the language-action outputs with respect to their grounding in the visual environment (see Section~\ref{subsec:method_clip}). 
Once the VLA and the grounding model are established, we proceed with grounded preference alignment by sampling multiple outputs from $\pi_\theta$, ranking them based on the $\pi_g$ scores, and aligning the VLA towards the highest-ranking outputs (see Section~\ref{subsec:preference_align}).
The GPLA training loop, based on Zhang et al.~\cite{zhang2024grapegeneralizingrobotpolicy}, is described in Algorithm~\ref{alg:training}.

\subsection{Dataset}
We use the LanguageTable benchmark suite~\cite{lynchInteractiveLanguageTalking2022} for its clear annotation into hierarchical steps to complete a given high-level task. 
An episode in the dataset displays a Franka Research Robot arm\footnote{\href{https://franka.de/franka-research-3}{https://franka.de/franka-research-3}} pushing blocks into position until the abstract high-level goal like "put all the blocks in a vertical line" is reached.
Segments of the episode are associated with a natural language description, describing what the robot's current action is. 
These captions, like "move the green circle near to the green star", directly represent how a human would break the high-level goal down into shorter executable low-level tasks.
As such, we use the captions as the target low-level instructions.
Regarding video and action data, a single frame of the dataset is taken from an angled top-down perspective, while actions are represented as a series of 2D-coordinates following the pointer at the end of the robot's arm.
During preprocessing, idle actions are removed by requiring the joint coordinate transposition in a given sample to surpass a certain threshold.
We empirically found a threshold of 0.1 in any action dimension to be effective in removing idle actions while keeping relevant actions.
When training our models, we apply data augmentation, including stochastic brightness/contrast adjustments, mild cropping, scaling on the frames, and Gaussian noise on the actions, ensuring transformations preserved task semantics (e.g., no left/right mirroring).

\subsection{Vision-Language-Action Model}\label{subsec:method_vla}
At their core, Vision-Language-Action models consist of a Vision-Language model backbone and a detokenization mechanism to map the vision-language output to executable policy actions. 
Prior work found it useful to break complex \textit{high-level instructions} down into simpler subgoals, which we will refer to as \textit{low-level instructions}, before passing them to the VLA. 
This can be done by using the same model which was co-trained on low-level instruction generation~\cite{blackpi05VisionLanguageActionModel2025,belkhaleRTHActionHierarchies2024} or by using a separate high-level module~\cite{shiHiRobotOpenEnded2025}.

\paragraph{Model Architecture}
To create our hierarchical VLA, we fine-tune a pre-trained Vision-Language-Action model and a pre-trained Vision-Language model. 
In the context of our hierarchical model, we will refer to the fine-tuned VLA as the \textit{low-level VLA} and the fine-tuned VLM as the \textit{high-level VLM}.
We choose a small Gemma3\footnote{Specifically Gemma-3-4B-IT}~\cite{teamGemma3Technical2025} VLM as our high-level VLM, which breaks the high-level instruction down into low-level instructions. 
We fine-tune the model to predict the low-level instructions provided by the LanguageTable dataset~\cite{lynchInteractiveLanguageTalking2022} from the high-level instruction.
During fine-tuning, the weights of the Gemma3 vision encoder are frozen.

For the low-level VLA, we independently fine-tune SmolVLA~\cite{shukorSmolVLAVisionLanguageActionModel2025a} on the ground-truth low-level instructions from the LanguageTable dataset for action generation.
During training, the low-level VLA learns to generate an 8-step trajectory conditioned on the image observation, end-effector state, high-level instruction, and ground-truth low-level instruction. 
At inference time, the ground-truth low-level instruction is replaced with the generated instruction of the high-level VLM. 

\subsection{Grounded Preference-based Language-Action Alignment}\label{subsec:preference_align}
After fine-tuning the VLA on generating low-level task descriptions alongside the trajectories, we iteratively improve the model using a preference-learning-based training scheme, similar to the one used by GRAPE~\cite{zhang2024grapegeneralizingrobotpolicy}.
First, we collect N different options for language-action pairs by prompting the model N times, given the same observation and instruction. 
We score the options with the grounding model and choose the option with the highest score as the preferred option and the one with the lowest as the rejected option.
Using these chosen/rejected preference pairs, we further train our model using SimPO~\cite{meng2024simpo}. 
We chose SimPO over DPO~\cite{rafailov2023dpo} as our training paradigm because SimPO does not rely on a reference model, which reduces the memory and computational requirements of our experiments while exhibiting similar performance. The SimPO loss for a single sample is defined as:
\begin{align}
    \mathcal{L}_{\text{SimPO}}(x, y_w, y_l) &= - \log \sigma \big( r(x,y_c) - r(x,y_r) - \gamma_{\text{SimPO}} \big)\\
    r(x,y) &= \frac{\beta_{\text{SimPO}}}{|y|} \log \pi_g(y|x)
\end{align}
With $y_{\text{c}}$ and $y_{\text{r}}$ as the highest and lowest scoring model outputs based on the grounding  score assignments, $\sigma$ is the sigmoid function
$\beta_{\text{SimPO}}$ and $\gamma_{\text{SimPO}}$ are tunable hyperparameters, and $r(x, y)$ represents the reward score of a response $y$ given a prompt $x$.
Using the SimPO loss, we directly update the high-level VLM of the hierarchical VLA.

\subsection{Action-Conditioned Grounding Model}\label{subsec:method_clip}
Hierarchical VLAs are typically trained with independent objectives for language and action outputs, leading to weak cross-modal alignment.
Evaluation mainly considers task success rates, ignoring the correctness of sub-steps.
However, ensuring a hierarchical VLA’s language output is grounded and aligned with the action space is challenging, as there is no established metric for quantifying this alignment.
Established contrastive models like SigLIP~\cite{tschannenSigLIP2Multilingual2025,zhaiSigmoidLossLanguage2023} or CLIP~\cite{radfordLearningTransferableVisual2021} can be used to score semantic alignment between vision and language modalities but don't incorporate an action modality.
Therefore, we separately train a grounding model to map vision-action pairs and language into a shared, aligned embedding space.

\begin{figure}[ht]
    \centering
    \includegraphics[trim={2em 1.5em 2em 2em},clip,width=\linewidth]{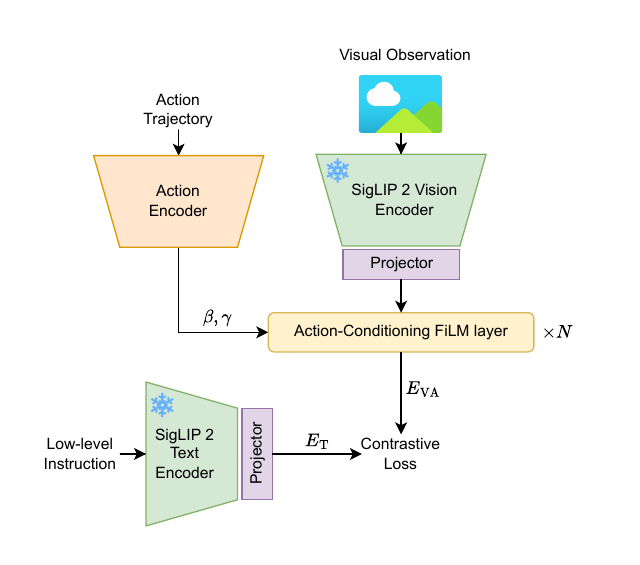}
    \caption{\textbf{Action-Conditioned Grounding Model.} We extend a pre-trained SigLIP 2 by conditioning the visual features of the SigLIP 2 Vision Encoder on the encoded trajectories. Using a contrastive loss, we align the vision-action pairs with the low-level instructions.}
    \label{fig:grounding-model}
\end{figure}

\paragraph{Model Architecture}
We train the action-conditioned grounding model to associate text descriptions with vision-action inputs. 
Its architecture is depicted in Figure~\ref{fig:grounding-model}.
Two separate encoder pipelines transform the inputs into a joint embedding space and align the latent representations using a symmetric InfoNCE loss function, as in CLIP~\cite{radfordLearningTransferableVisual2021}.
The vision and text encoders are initialized from a pre-trained SigLIP 2 model\footnote{Specifically, we use SigLIP 2 ViT B/16.}~\cite{tschannenSigLIP2Multilingual2025} and kept frozen during training.
We follow each pre-trained encoder up with a projection layer to further reduce the dimensionality of the pooled feature outputs.
The action encoder is a small transformer network.
We condition the projected vision features from the SigLIP 2 vision encoder on the embedded actions through a series of FiLM layers~\cite{perezFiLMVisualReasoning2018} that allow the visual features to be dynamically modulated based on the action context.
Finally, the loss is calculated on the embedded action-vision and language representations.
The goal is to ensure that correctly aligned vision-action and text inputs yield closely-matched embeddings.

\paragraph{Objective Function}
We use the symmetric InfoNCE loss as in CLIP~\cite{radfordLearningTransferableVisual2021} as the objective function to align the vision-action embedding $E_{\text{VA}}$ with the text embedding $E_\text{T}$. 
Before calculating the loss, we normalize the embeddings and refer to the normalized embeddings as $\text{VA}$ and $\text{T}$.
The loss is defined as:
\begin{align}\label{eq:contrastive_loss}
    L_{\text{VA} \to \text{T}} &= -\sum_{i=1}^{N} \log \frac{\exp(sim(\text{VA}_i, \text{T}_i) / \tau)}{\sum_{k=1}^{N} \exp(sim(\text{VA}_i, \text{T}_k) / \tau)} \\
    L_{\text{T} \to \text{VA}} &= -\sum_{j=1}^{N} \log \frac{\exp(sim(\text{VA}_j, \text{T}_j) / \tau)}{\sum_{k=1}^{N} \exp(sim(\text{VA}_k, \text{T}_j) / \tau)}\\
    L_C &= \frac{1}{2} (L_{\text{VA} \to \text{T}} + L_{\text{T} \to \text{VA}})
\end{align}
In equations \ref{eq:contrastive_loss}, $\tau$ is a learned temperature parameter that scales the logits before the softmax and $sim$ is the cosine similarity function.

Additionally, to prevent the model from collapsing all embeddings into a single representation and instead encourage learning separable embeddings, we additionally introduce a diversity regularization term $L_{div}$, which we add to the contrastive loss:
\begin{align}\label{eq:diversity}
    L_{\text{div}} &= \frac{1}{N(N-1)} \sum_{i \neq j} \bigl( \max(0, S_{\text{VA}}^{ij}) + \max(0, S_{\text{T}}^{ij}) \bigr) \\
    L &= L_C + \gamma_{{\text{div}}} L_{\text{div}}
\end{align}
where $S=EE^T$ is the cosine similarity matrix for the respective embeddings.
The hyperparameter $\gamma_{\text{div}}$ controls the influence of the regularization term. 
Intuitively, the diversity regularization term encourages the model to have low similarity values along the off-diagonal elements of the similarity matrix. 

%% file: sec/5_experiments.tex
\begin{table*}[ht]
\centering
\scriptsize 
\caption{Comparison of model variants on the language- and trajectory-based metrics.}
\begin{tabular}{lccccccc}
\toprule
\textbf{Model} 
& \textbf{BLEU}$\uparrow$ 
& \textbf{ROUGE}$\uparrow$ 
& \textbf{METEOR}$\uparrow$ 
& \textbf{BERTScore}$\uparrow$ 
& \textbf{MSE}$\downarrow$ 
& \textbf{MAE}$\downarrow$  
& \textbf{CosSim}$\downarrow$  \\
\midrule
Low-Level Only & N/A & N/A & N/A & N/A & 0.043 $\pm$ 0.02 & 0.158 $\pm$ 0.04 & -0.029 $\pm$ 0.22 \\
Supervised & 0.111 $\pm$ 0.05 & 0.405 $\pm$ 0.12 & 0.313 $\pm$ 0.12 & 0.984 $\pm$ 0.00 & 0.046 $\pm$ 0.02 & 0.164 $\pm$ 0.04 & -0.044 $\pm$ 0.23 \\
GPLA (CLIP) & 0.062 $\pm$ 0.04 & 0.298 $\pm$ 0.12 & 0.217 $\pm$ 0.12 &  0.976 $\pm$ 0.00  & 0.045 $\pm$ 0.02  & 0.164 $\pm$ 0.04 & -0.036 $\pm$ 0.22 \\
GPLA (SigLIP 2) & 0.066 $\pm$  0.06 & 0.307 $\pm$ 0.13 & 0.227 $\pm$ 0.12 & 0.976 $\pm$ 0.00 & 0.045 $\pm$ 0.02  & 0.163 $\pm$ 0.04 & -0.035 $\pm$ 0.22\\
GPLA (Action-Conditioned) & 0.063 $\pm$ 0.05 & 0.300 $\pm$ 0.12 & 0.218 $\pm$ 0.12 & 0.980 $\pm$ 0.00  & 0.045 $\pm$ 0.02  & 0.163 $\pm$ 0.04 & -0.035 $\pm$ 0.22\\
Supervised + GPLA (Action-Conditioned) & 0.051 $\pm$ 0.05 & 0.308 $\pm$ 0.12 & 0.226 $\pm$ 0.12 & 0.980 $\pm$ 0.00 & 0.046 $\pm$ 0.02  & 0.163 $\pm$ 0.04 & -0.042 $\pm$ 0.23\\
\bottomrule
\end{tabular}
\label{tab:combined_model_comparison}
\end{table*}
\section{Experimental Setup}
We train all our models on a single NVIDIA A100 GPU. 
The hierarchical VLA is trained to generate actions with a horizon of 8; the grounding model learns to associate trajectories (horizon of 8) and visual inputs with low-level instructions.
Training times differ between approaches:
To establish the baseline hierarchical VLA, the supervised fine-tuning of the high-level VLM takes 1,500 steps with a learning rate of $10^{-5}$, the AdamW Optimizer and no scheduler; the low-level VLA is trained for 15,000 steps also using a fixed learning rate of $10^{-5}$ and the AdamW optimizer without scheduling. Both operate on batches of size of 64. 
We then apply our framework on the established baseline hierarchical VLA for an additional 100 steps, with a learning rate of $10^{-7}$.
The grounding model is trained for 50,000 steps with a fixed learning rate of $10^{-4}$ and a larger effective batch size of 256 after accumulating batches of size 64 for 4 forward-passes.

%% file: sec/6_results.tex
\section{Results}
We investigate the effectiveness of GPLA on the capability to follow instructions on episodes of the LanguageTable dataset that were withheld during training. 
We analyze the model's ability to improve grounding its high-level statements to the low-level actions by investigating the statement correctness using text-based metrics, like BLEU~\cite{papipeni2002bleu}, ROUGE\footnote{Specifically, ROUGE-1 $\text{F}_1$-measure.}~\cite{lin-2004-rouge}, METEOR~\cite{banerjee2005meteor}, and BERTScore~\cite{zhang2020BERTScore}.
We report instruction following capabilities based on the deviations from the ground truth trajectories, given as MAE~\cite{tervenComprehensiveSurveyLoss2025}, MSE~\cite{tervenComprehensiveSurveyLoss2025}, and cosine similarity~\cite{tervenComprehensiveSurveyLoss2025}.
The detailed quantitative results are summarized in Table~\ref{tab:combined_model_comparison}.

\begin{table*}[!ht]
\centering
\footnotesize 
\renewcommand{\arraystretch}{1.3}
\setlength{\tabcolsep}{3pt} 

\caption{Qualitative Examples on the LanguageTable Dataset~\cite{lynchInteractiveLanguageTalking2022}.}
\label{tab:qualitative_examples}
\begin{tabular}{
    >{\RaggedRight\arraybackslash}m{3.4cm}  
    >{\centering\arraybackslash}p{3.1cm}    
    >{\centering\arraybackslash}p{3.1cm}
    >{\centering\arraybackslash}p{3.1cm}
    >{\centering\arraybackslash}p{3.1cm}
}
\toprule
  & \includegraphics[width=0.18\textwidth]{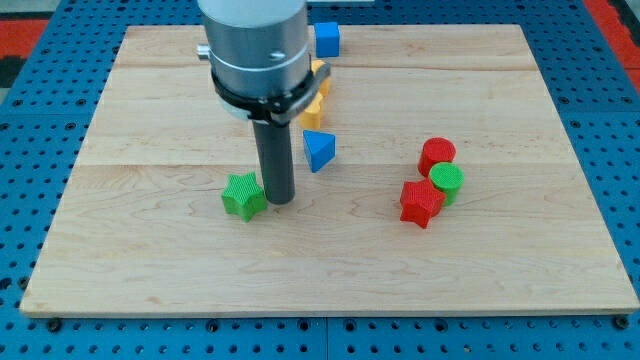}
    & \includegraphics[width=0.18\textwidth]{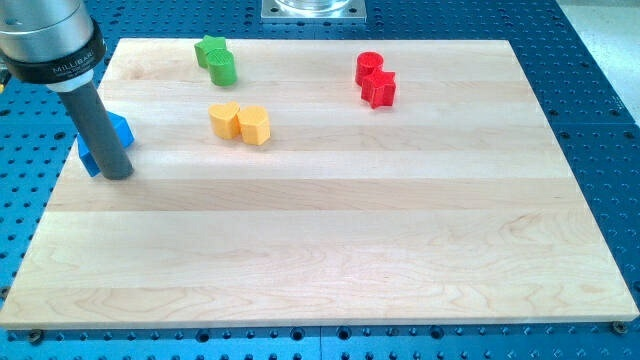}
    & \includegraphics[width=0.18\textwidth]{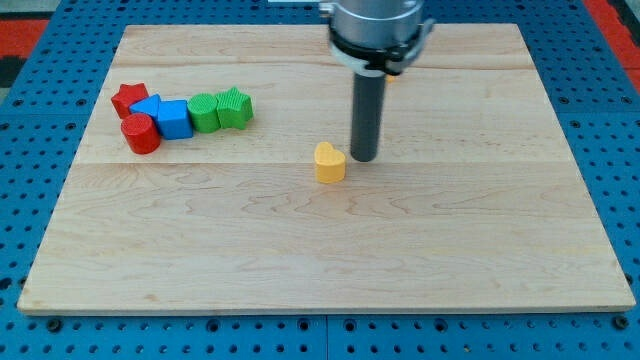}
    & \includegraphics[width=0.18\textwidth]{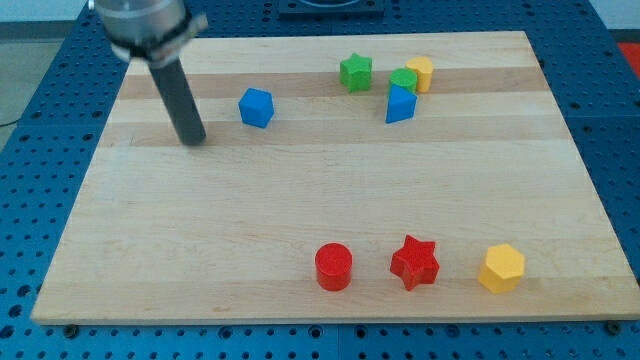} \\

\midrule
High-level Instructions &
make a "parallelogram" shape out of all the blocks &
put all the blocks in the bottom left corner &
put all the blocks in the center left 
 & put all the blocks in a horizontal line on the bottom of the board
 \\

Low-level Instructions (GT) &
move the green star diagonal to the hexagon &
move the blue blocks towards the bottom left &
keep the yellow heart at the bottom right side of the green star
 & move the blue blocks towards the bottom left
 \\

\midrule
\textcolor{plotblue}{\textbf{Supervised}} & \textcolor{plotblue}{move your arm towards the left below the yellow heart}
 & \textcolor{plotblue}{push the yellow hexagon into your hand}
 & \textcolor{plotblue}{set down the heart}
 & \textcolor{plotblue}{slide the blue cube slightly towards the left and the right of the blue triangle}  \\
\textcolor{plotorange}{\textbf{GPLA (Action-Conditioned)}} & \textcolor{plotorange}{place hexagon above square} & \textcolor{plotorange}{move your arm towards yellow star} & \textcolor{plotorange}{move your arm towards front of the board} & \textcolor{plotorange}{place your arm towards towards left side} \\
\textcolor{plotgreen}{\textbf{Supervised + GPLA (Action-Conditioned)}} & \textcolor{plotgreen}{push the red circle diagonally to the triangle} & \textcolor{plotgreen}{move yellow hexagon into red star} & \textcolor{plotgreen}{drag red circle to the yellow hexagon} & \textcolor{plotgreen}{push blue cube diagonally above green circle } \\
\bottomrule
\end{tabular}

\end{table*}

\subsection{Low-level Instruction Generation}\label{subsec:instruction_generation}
\begin{figure}
    \centering
    \includegraphics[width=\linewidth]{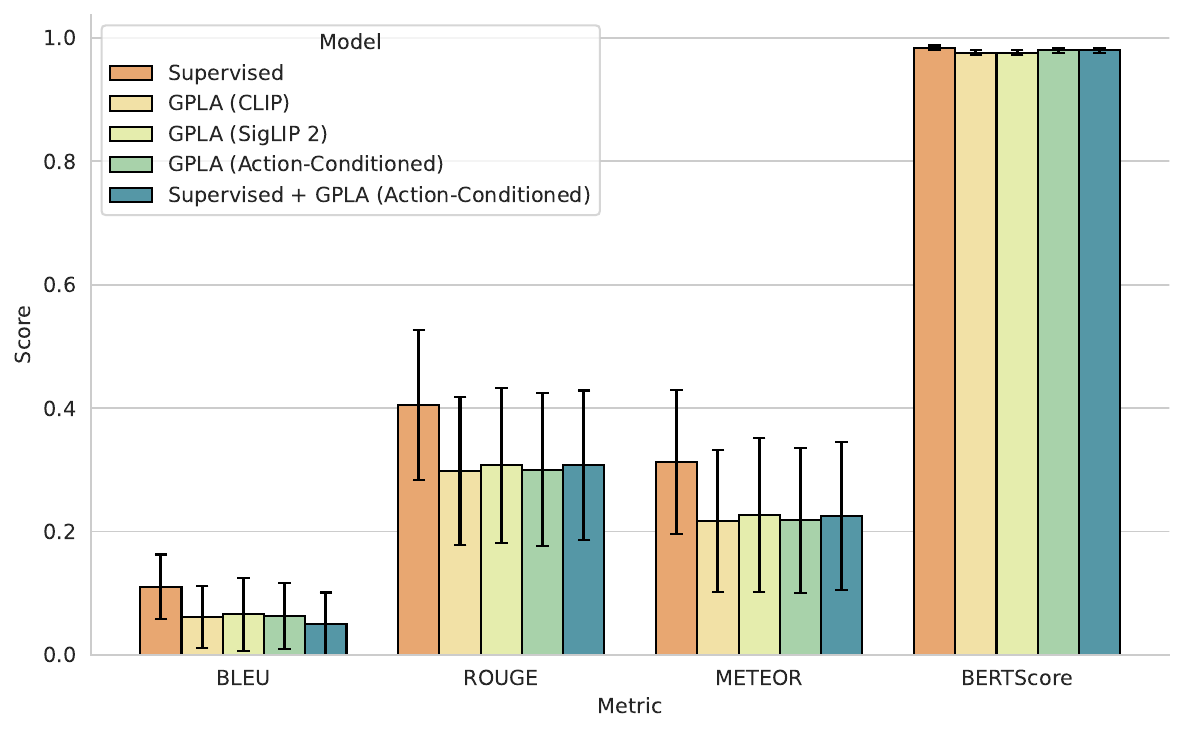}
    \caption{Quantitative Evaluation of Generated Low-level Instructions.}
    \label{fig:language_metrics_plot}
\end{figure}

We investigate the impact of the novel GPLA framework on the low-level instruction generation within our hierarchical VLA.
While the quantitative results in Figure~\ref{fig:language_metrics_plot} show a decrease in token-overlap metrics (BLEU, ROUGE, and METEOR) compared to purely supervised training, this shift is achieved without requiring any additional low-level ground-truth data.
Critically, the semantic score (BERTScore) remains stable, suggesting that the model keeps generating semantically coherent outputs which are not necessary captured by token-based measures.
Based on the notion that supervised learning remains a crucial component of the training recipe, we incorporate the GPLA objective as a weighted regularization term with a weight of 0.1 into the standard language modeling loss. 
This change yields mixed results compared to pure supervision or preference-based grounding. 
Except for the BLEU metrics, ROGUE, BERTScore, and METEOR score display slight increases.
Using SigLIP 2 as a grounding model fares better than CLIP and our action-conditioned grounding model, and compares to using the GPLA with the action-conditioned grounding model as a regularization term. 
The qualitative examples in Table~\ref{tab:qualitative_examples} provide crucial context for interpreting the token-based metrics (e.g., BLEU, ROUGE, and METEOR).
While these metrics yield comparatively low values due to limited word-overlap with the low-level ground-truth instructions, all three models consistently produce generally compelling-sounding instructions.
A detailed analysis highlights that object relations are a strong pathway for generating subsequent commands. 
In fact, four out of the twelve analyzed examples featured a movement associated with the relative position of at least one other block.
Moreover, the models successfully identified and targeted objects, including shape and color, across most instances; the key challenge lies in accurately determining the correct spatial relationship and ensuring the command is executable.
We observed two primary challenges that currently limit performance: First, the supervised training occasionally suggested actions that are physically incompatible with the agent's embodiment, such as “[...] into your hand” or “set down [...]” - a specific area for refining spatial and embodied grounding in current VLMs. 
Second, pure preference grounding introduced isolated instances of linguistic noise (e.g., “[...] towards towards [...]”). 
Addressing these specific challenges will enable the models to fully leverage their strong object-relation and command-generation capabilities.

\begin{figure}
    \centering
    \includegraphics[width=\linewidth]{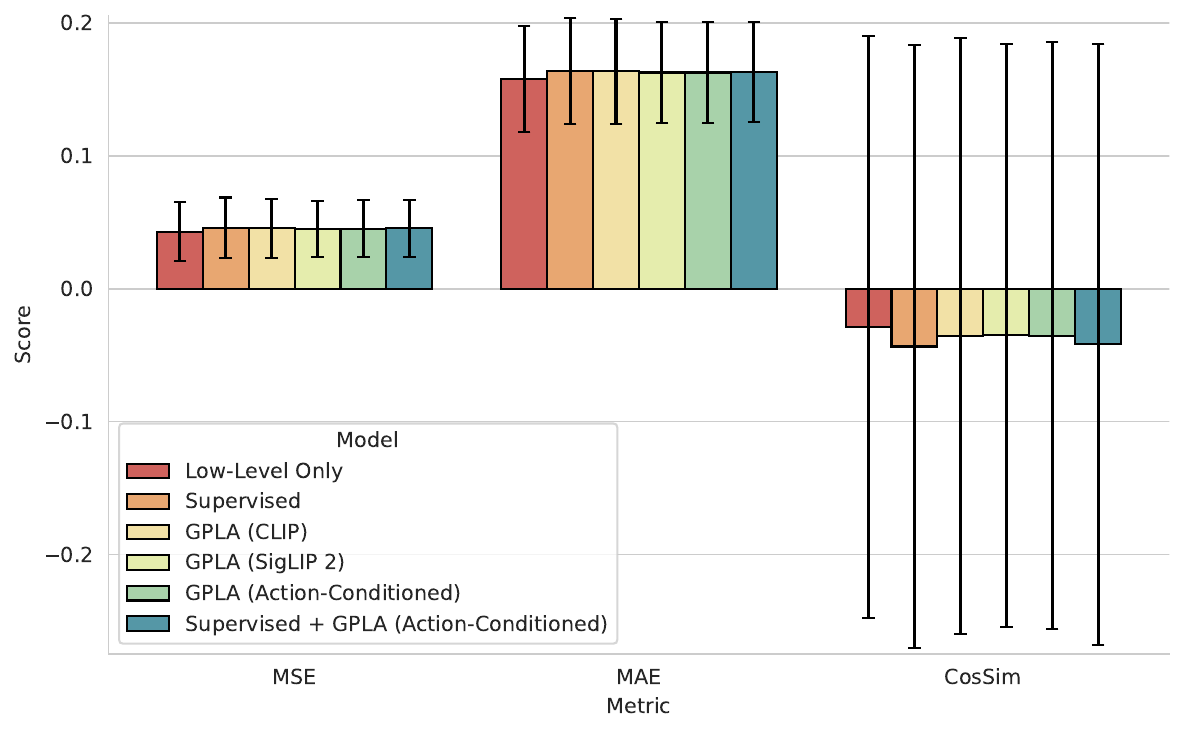}
    \caption{Quantitative Evaluation of Generated Trajectories.}
    \label{fig:action_metrics_plot}
\end{figure}

\begin{figure*}[ht]
    \centering
    
    \begin{subfigure}[t]{0.32\textwidth}
        \centering
        \includegraphics[width=\textwidth]{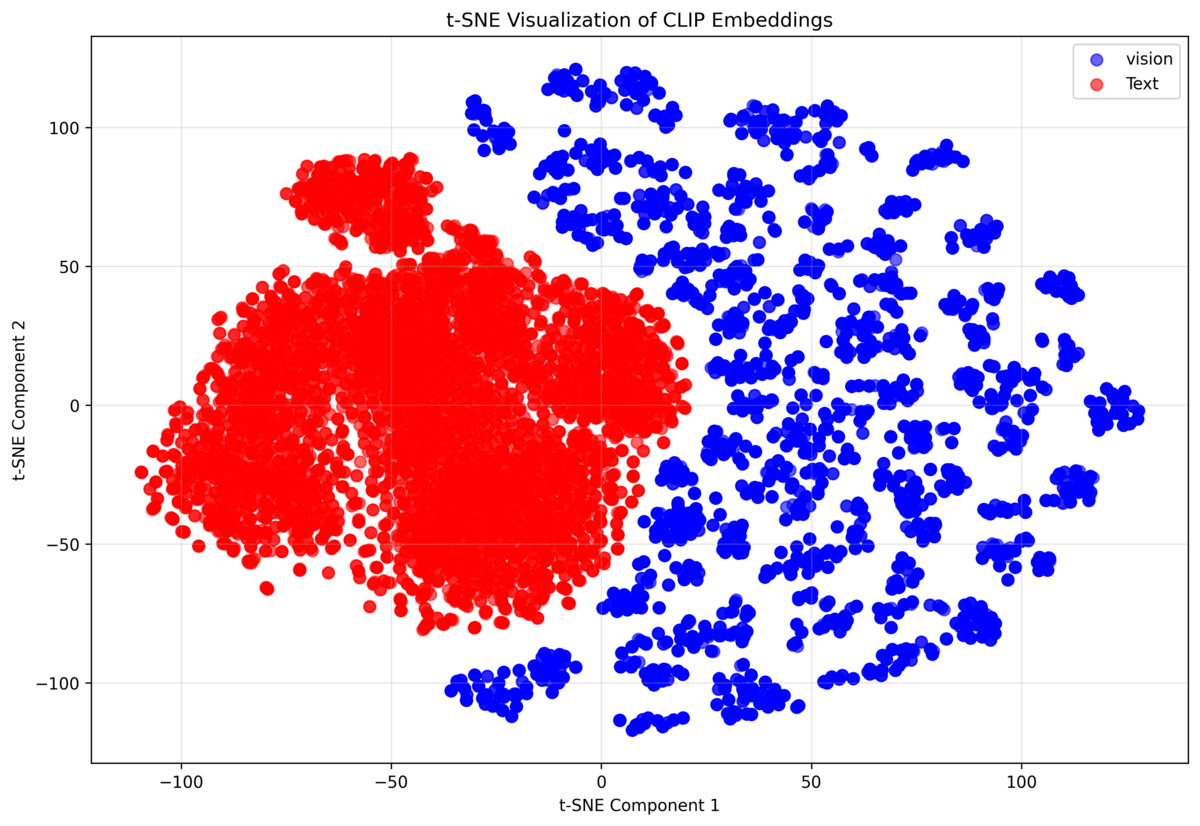}
        \caption{t-SNE visualization of CLIP embeddings.}
        \label{fig:tsne_clip}
    \end{subfigure}
    \hfill
    \begin{subfigure}[t]{0.32\textwidth}
        \centering
        \includegraphics[width=\textwidth]{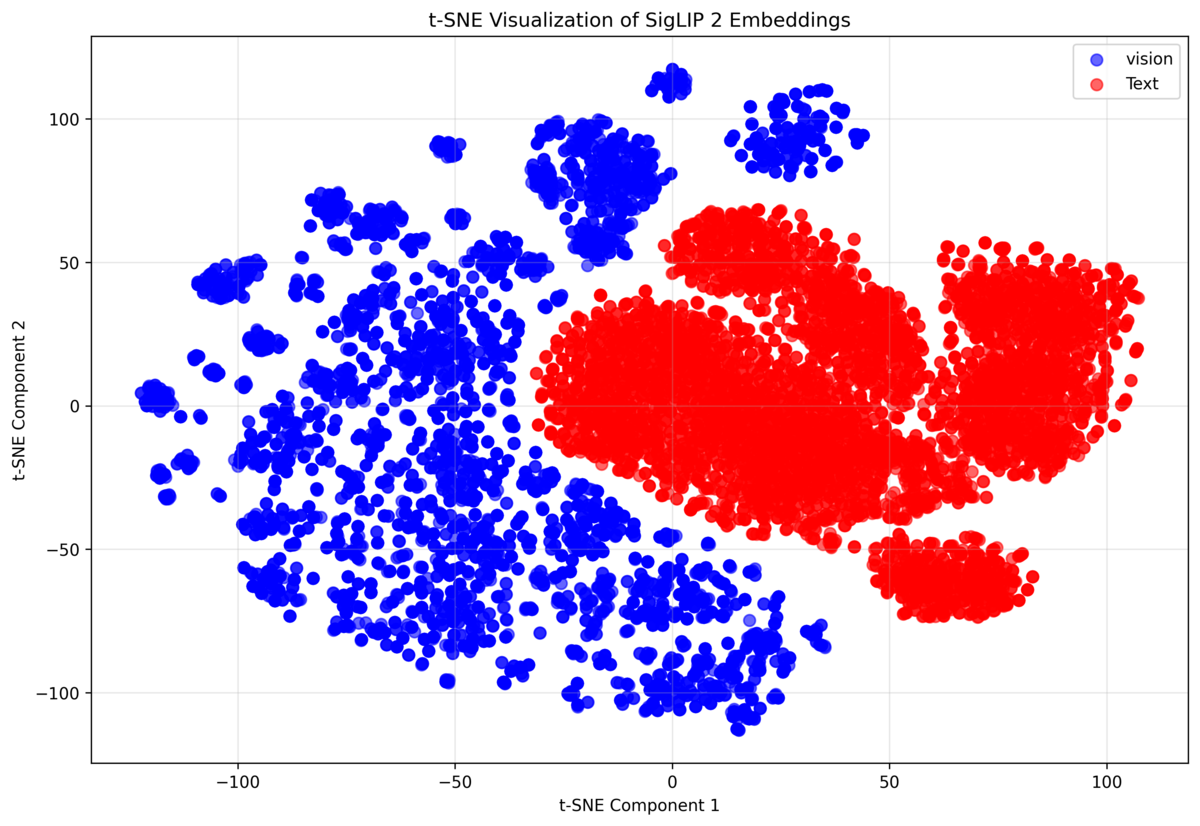}
        \caption{t-SNE visualization of SigLIP 2 embeddings.}
        \label{fig:tsne_siglip}
    \end{subfigure}
    \hfill
    \begin{subfigure}[t]{0.32\textwidth}
        \centering
        \includegraphics[width=\textwidth]{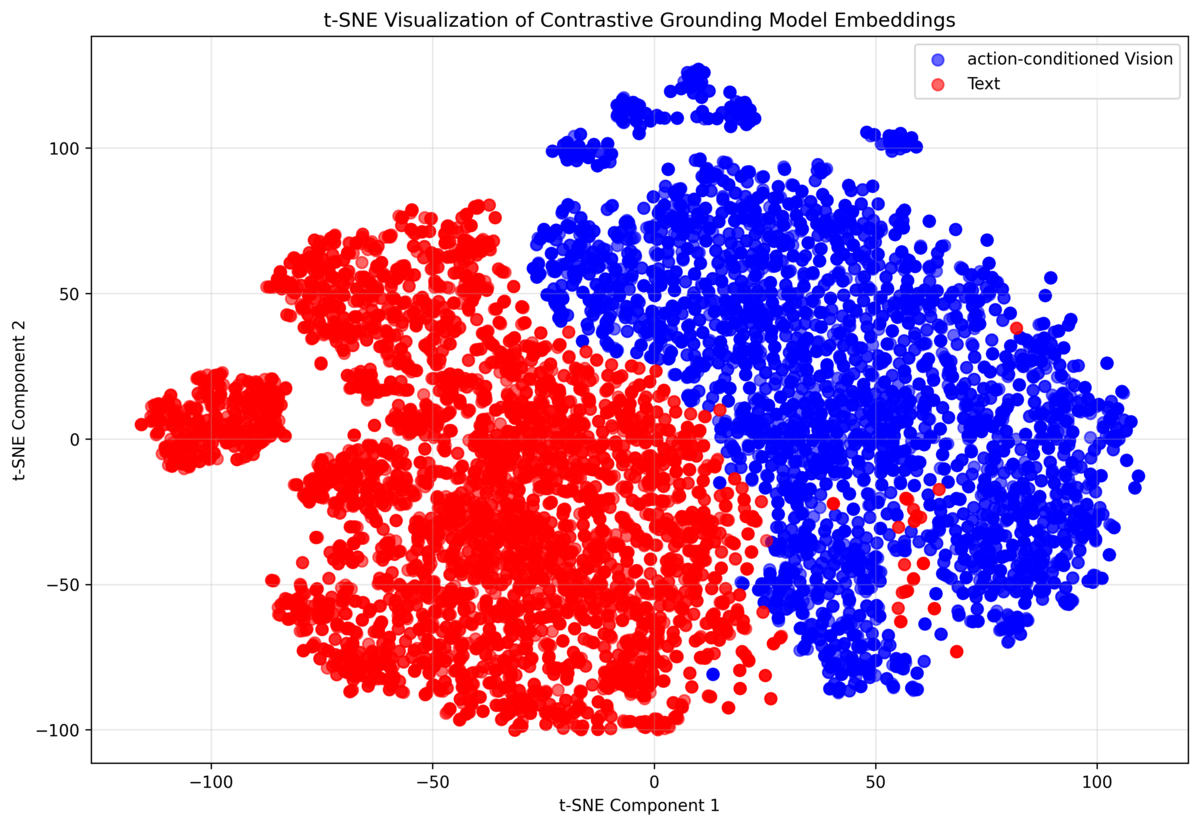}
        \caption{t-SNE visualization of action-conditioned grounding model embeddings.}
        \label{fig:tsne_ours}
    \end{subfigure}
    
    \caption{t-SNE visualizations of all grounding models on the LanguageTable dataset using visual inputs and low-level instructions.}
    \label{fig:tsne_all}
\end{figure*}
\subsection{Trajectory Generation}\label{subsec:traj_gen}
Considering the generated trajectories, we compare our model variants additionally to the supervised SmolVLA, which receives the ground-truth low-level instructions. 
While the supervised model performs best given the ground-truth instructions, most GPLA versions exhibit similar performance. 
The performance of the VLA remains robust across all GPLA variants, evidenced by both the magnitude-sensitive metrics (MSE and MAE) and the directional metric, cosine similarity.
Among these metrics, cosine similarity demonstrates the best performance for pure GPLA variations.
Collectively, these findings highlight that the high-level VLM provides the low-level VLA with semantically useful instructions.

\subsection{Contrastive Embedding space}
To gain further insight into the applicability of different contrastive models for the purpose of grounding language in vision-action outputs, Figure~\ref{fig:tsne_all} depicts the t-SNE visualizations with a perplexity of 30 for the pre-trained CLIP and SigLIP 2 models, as well as our action-conditioned grounding model.
All models show signs of visually separable clusters in the vision and text domains to varying degrees. 
As depicted in Figure~\ref{fig:tsne_clip}, CLIP’s embedding space shows a well-defined separation across visual inputs, indicating strong visual discrimination. 
The language embeddings, by comparison, mostly form a single cohesive cluster with a few internal substructures and one clearly separated group. 
The only model exhibiting signs of mixing vision and language embeddings is the action-conditioned grounding model. This is evidenced by a small language sub-structure appearing within the main vision-cluster (see lower right corner of Figure~\ref{fig:tsne_ours}). 
This blending suggests that the action-conditioning encourages the model to learn a unified, multi-modal representation where linguistic and visual elements of the current step are fused, unlike the distinct separation seen in the other models.  

%% file: sec/7_discussion.tex
\section{Discussion}
The GPLA framework offers a compelling solution for data-scarce scenarios. 
While supervised training methods achieve optimal results, they necessitate costly amounts of annotated data. 
Crucially, the method presented in this paper validates that a preference-based approach can successfully retain comparable performance levels (see Section~\ref{subsec:traj_gen}) without relying on expensive expert annotation.
Supervised learning inherently struggles to deal with semantic ambiguities, often failing to adequately capture the correctness of acceptable, yet diverse, outputs within its rigid loss function. 
Preference-based methods directly address this difficulty by leveraging comparative feedback, which is ideally suited to resolve these subtle semantic differences.

Despite the successful application of foundation model-backed VLAs, further work is required to improve their understanding of object relations and embodiment, a difficulty explored in Section~\ref{subsec:instruction_generation}.
Even common pre-trained models are largely incapable of inferring grounding directions, colors, and object relations, and infer reasonable steps from the task instructions beyond the recognition of different shapes~\cite{cheng2024spatialrgpt}.
We continue to hypothesize that VLA and VLM reasoning and planning capabilities need to incorporate stronger grounding mechanisms into their training paradigm, like those established by our proposed framework, GPLA. 

Regarding the grounding model, contrastive learning has shown to be most effective with vast amounts of data and large batch sizes. 
Given the necessary constraint of a relatively small corpus of diverse, language-annotated robotic episodes, GPLA successfully validates the preference-based grounding approach. 
We hypothesize that, while current performance is constrained by data scarcity and available compute, our method is ideally positioned to leverage future large-scale multimodal robotic datasets for substantial performance gains.



%% file: sec/8_conclusion.tex
\section{Conclusion}
Grounding natural language in real-world actions remains a challenging task. 
We propose a preference-based framework, GPLA, to improve hierarchical VLAs, which refine a high-level instruction into a low-level instruction before generating a trajectory. 
Our framework explicitly uses a learned grounding model to ensure the generation of the low-level instruction aligns with the action-trajectory. 
Applying our novel preference-based framework to the LanguageTable benchmark, we establish a strong baseline with performance comparable to fully supervised fine-tuning, critically demonstrating that high-quality action-language grounding can be maintained without the need for costly, extensive data annotation, while simultaneously delivering crucial insights into multimodal grounding representations.

\section*{Acknowledgements}
The authors gratefully acknowledge funding from the EU and UKRI in the context of Horizon Europe under the MSCA grant agreement No 101072488 (TRAIL). Special thanks also the team at the Computational Shared Facility at the University of Manchester for providing the resources to train our models.

%% file: sec/9_appendix.tex
\appendix
\section{Prompt}
We use a conversational structure to prompt the high-level VLM for low-level instructions. Figure~\ref{fig:prompt_template} depicts the corresponding prompt template. 
The answer not provided during inference.
\begin{figure}[ht]
\centering
\fbox{%
    \parbox{0.95\linewidth}{%
\ttfamily
System: You are controlling a robotic agent. Your task is to <high-level instruction>.\\
User: What should the robot do next?\\
Answer: <low-level instruction>
    }%
}
\caption{Prompt template used for robotic agent instructions.}
\label{fig:prompt_template}
\end{figure}
\section{Data Augmentation}
The data augmentation techniques we applied during training are listed in Table~\ref{tab:augmentations}.
\begin{table}[ht]
\centering
\begin{tabular}{llc}
\hline
\textbf{Modality} & \textbf{Augmentation} & \textbf{Probability} \\
\hline
\multirow{7}{*}{Images} 
 & Brightness          & 0.5 \\
 & Contrast            & 0.5 \\
 & Saturation          & 0.5 \\
 & Crop and resize     & 0.6 \\
 & Vertical translation & 0.4 \\
 & Horizontal translation & 0.4 \\
 & Scale (zoom in/out) & 0.3 \\
\hline
\multirow{1}{*}{Actions} 
 & Noise & 0.7 \\
\hline
\end{tabular}
\caption{Data augmentation techniques and their application probabilities, grouped by modality.}
\label{tab:augmentations}
\end{table}
\newpage
\section{Hyperparameters}
Table~\ref{tab:hyperparameters} lists the hyperparameters we applied to during training on the different model variants.\\
\begin{table}[ht]
\small
\centering
\begin{tabular}{ll}
\hline
\multicolumn{2}{c}{\textbf{General}} \\
\hline
Gradient norm clipping: & 1.0 \\
Action cut-off threshold: & 0.1 \\
\hline
\multicolumn{2}{c}{\textbf{High-level VLM (fine-tuning)}} \\
\hline
Steps: & 1{,}500 \\
Learning rate: & $10^{-5}$ \\
Effective Batch size: & 64 \\
Optimizer: & AdamW \\
Horizon: & 8 \\
\hline
\multicolumn{2}{c}{\textbf{Low-level VLA (fine-tuning)}} \\
\hline
Steps: & 15{,}000 \\
Learning rate: & $10^{-5}$ \\
Effective Batch size: & 64 \\
Optimizer: & AdamW \\
Horizon: & 8 \\
\hline
\multicolumn{2}{c}{\textbf{GPLA}}\\
\hline
Steps: & 100 \\
Learning rate: & $10^{-7}$ \\
Batch size: & 64 \\
Optimizer: & AdamW \\
Horizon: & 8 \\
\hline
\multicolumn{2}{c}{\textbf{Action-Conditioned Grounding Model}} \\
\hline
Steps: & 50{,}000 \\
Learning rate: & $10^{-4}$ \\
Effective batch size: & 256 \\
Optimizer: & Adam \\
Horizon: & 8 \\
Initial logit scale factor: & 0.1 \\
Label smoothing: & None \\
Diversity weight: & 0.01 \\
Model dimension: & 64 \\
N\_FiLM layers: & 4 \\
\hline
\end{tabular}
\caption{Hyperparameters grouped by model variant.}
\label{tab:hyperparameters}
\end{table}

%% file: main.bib
@string(CVPR= {IEEE Conf. Comput. Vis. Pattern Recog.})

@string(ICCV= {Int. Conf. Comput. Vis.})

@string(ICLR = {Int. Conf. Learn. Represent.})

@string(AAAI = {AAAI})

@string(CVPR  = {CVPR})

@string(ICCV  = {ICCV})

@string(ICLR  = {ICLR})

@inproceedings{ma2023liv,
	title        = {{LIV}: {Language-Image Representations and Rewards for Robotic Control}},
	author       = {Yecheng Jason Ma and Vikash Kumar and Amy Zhang and Osbert Bastani and Dinesh Jayaraman},
	year         = 2023,
	booktitle    = {Workshop on Reincarnating Reinforcement Learning at ICLR 2023},
	url          = {https://openreview.net/forum?id=DtsUiQwEcE}
}

@inproceedings{sontakke2023roboclip,
	title        = {Robo{CLIP}: {One Demonstration is Enough to Learn Robot Policies}},
	author       = {Sumedh Anand Sontakke and Jesse Zhang and S{\'e}b Arnold and Karl Pertsch and Erdem Biyik and Dorsa Sadigh and Chelsea Finn and Laurent Itti},
	year         = 2023,
	booktitle    = {Thirty-seventh Conference on Neural Information Processing Systems},
	url          = {https://openreview.net/forum?id=DVlawv2rSI}
}

@article{weiChainofThoughtPromptingElicits2022,
	title        = {Chain-of-{{Thought Prompting Elicits Reasoning}} in {{Large Language Models}}},
	author       = {Wei, Jason and Wang, Xuezhi and Schuurmans, Dale and Bosma, Maarten and Ichter, Brian and Xia, Fei and Chi, Ed and Le, Quoc V. and Zhou, Denny},
	year         = 2022,
	month        = dec,
	journal      = {Advances in Neural Information Processing Systems},
	volume       = 35,
	pages        = {24824--24837},
	urldate      = {2023-11-08},
	langid       = {english}
}

@inproceedings{ichterCanNotSay2023,
	title        = {Do {{As I Can}}, {{Not As I Say}}: {{Grounding Language}} in {{Robotic Affordances}}},
	shorttitle   = {Do {{As I Can}}, {{Not As I Say}}},
	author       = {Ichter, Brian and Brohan, Anthony and Chebotar, Yevgen and Finn, Chelsea and Hausman, Karol and Herzog, Alexander and Ho, Daniel and Ibarz, Julian and Irpan, Alex and Jang, Eric and Julian, Ryan and Kalashnikov, Dmitry and Levine, Sergey and Lu, Yao and Parada, Carolina and Rao, Kanishka and Sermanet, Pierre and Toshev, Alexander T. and Vanhoucke, Vincent and Xia, Fei and Xiao, Ted and Xu, Peng and Yan, Mengyuan and Brown, Noah and Ahn, Michael and Cortes, Omar and Sievers, Nicolas and Tan, Clayton and Xu, Sichun and Reyes, Diego and Rettinghouse, Jarek and Quiambao, Jornell and Pastor, Peter and Luu, Linda and Lee, Kuang-Huei and Kuang, Yuheng and Jesmonth, Sally and Joshi, Nikhil J. and Jeffrey, Kyle and Ruano, Rosario Jauregui and Hsu, Jasmine and Gopalakrishnan, Keerthana and David, Byron and Zeng, Andy and Fu, Chuyuan Kelly},
	year         = 2023,
	month        = mar,
	booktitle    = {Proceedings of {{The}} 6th {{Conference}} on {{Robot Learning}}},
	publisher    = {PMLR},
	pages        = {287--318},
	issn         = {2640-3498},
	urldate      = {2023-09-26},
	langid       = {english}
}

@article{harnadSymbolGroundingProblem1990,
	title        = {{The Symbol Grounding Problem}},
	author       = {Harnad, Stevan},
	year         = 1990,
	month        = jun,
	journal      = {Physica D: Nonlinear Phenomena},
	volume       = 42,
	number       = 1,
	pages        = {335--346},
	doi          = {10.1016/0167-2789(90)90087-6},
	issn         = {0167-2789},
	urldate      = {2025-08-26}
}

@inproceedings{stiennonLearningSummarizeHuman2020,
	title        = {{Learning to Summarize with Human Feedback}},
	author       = {Stiennon, Nisan and Ouyang, Long and Wu, Jeffrey and Ziegler, Daniel and Lowe, Ryan and Voss, Chelsea and Radford, Alec and Amodei, Dario and Christiano, Paul F},
	year         = 2020,
	booktitle    = {Advances in {{Neural Information Processing Systems}}},
	publisher    = {Curran Associates, Inc.},
	volume       = 33,
	pages        = {3008--3021},
	urldate      = {2024-01-31}
}

@article{han2021VErbal,
	title        = {{The Need for Verbal Robot Explanations and How People Would Like a Robot to Explain Itself}},
	author       = {Han, Zhao and Phillips, Elizabeth and Yanco, Holly A.},
	year         = 2021,
	month        = sep,
	journal      = {J. Hum.-Robot Interact.},
	publisher    = {Association for Computing Machinery},
	address      = {New York, NY, USA},
	volume       = 10,
	number       = 4,
	doi          = {10.1145/3469652},
	url          = {https://doi.org/10.1145/3469652},
	issue_date   = {December 2021},
	articleno    = 36,
	numpages     = 42
}

@inproceedings{nair2022r3m,
	title        = {{R3M: A Universal Visual Representation for Robot Manipulation}},
	author       = {Suraj Nair and Aravind Rajeswaran and Vikash Kumar and Chelsea Finn and Abhinav Gupta},
	year         = 2022,
	booktitle    = {6th Annual Conference on Robot Learning (CoRL)},
	url          = {https://openreview.net/forum?id=tGbpgz6yOrI}
}

@inproceedings{iii2022fewshot,
    title        = {{Few-Shot Preference Learning for Human-in-the-Loop} {RL}},
	author       = {Donald Joseph Hejna III and Dorsa Sadigh},
	year         = 2022,
	booktitle    = {6th Annual Conference on Robot Learning},
	url          = {https://openreview.net/forum?id=IKC5TfXLuW0}
}

@misc{giannone2025supervisionfreevisionlanguagealignment,
      title={{Feedback-Driven Vision-Language Alignment with Minimal Human Supervision}}, 
      author={Giorgio Giannone and Ruoteng Li and Qianli Feng and Evgeny Perevodchikov and Rui Chen and Aleix Martinez},
      year={2025},
      eprint={2501.04568},
      archivePrefix={arXiv},
      primaryClass={cs.CV},
      url={https://arxiv.org/abs/2501.04568}, 
      note = {arXiv:2501.04568 [cs.CV]}
}

@inproceedings{shao2020concept,
	title        = {{Concept2Robot: Learning Manipulation Concepts from Instructions and Human Demonstrations}},
	author       = {Shao, Lin and Migimatsu, Toki and Zhang, Qiang and Yang, Karen and Bohg, Jeannette},
	year         = 2020,
	booktitle    = {Proceedings of Robotics: Science and Systems (RSS)}
}

@inproceedings{lee2021pebble,
	title        = {{PEBBLE: Feedback-Efficient Interactive Reinforcement Learning via Relabeling Experience and Unsupervised Pre-training}},
	author       = {Kimin Lee and Laura M. Smith and P. Abbeel},
	year         = 2021,
	booktitle    = {International Conference on Machine Learning},
	url          = {https://api.semanticscholar.org/CorpusID:235377145}
}

@INPROCEEDINGS{myers2023activerewardlearningonline,
  author={Myers, Vivek and Bıyık, Erdem and Sadigh, Dorsa},
  booktitle={2023 IEEE International Conference on Robotics and Automation (ICRA)}, 
  title={Active Reward Learning from Online Preferences}, 
  year={2023},
  volume={},
  number={},
  pages={7511-7518},
  doi={10.1109/ICRA48891.2023.10160439}}

@inproceedings{
    zhang2024grapegeneralizingrobotpolicy,
    title={{GRAPE}: Generalizing Robot Policy via Preference Alignment},
    author={Zijian Zhang and Kaiyuan Zheng and Zhaorun Chen and Joel Jang and Yi Li and Siwei Han and Chaoqi Wang and Mingyu Ding and Dieter Fox and Huaxiu Yao},
    booktitle={Workshop on Reasoning and Planning for Large Language Models},
    year={2025},
    url={https://openreview.net/forum?id=XnwyFD1Fvw}
}

@misc{ma2023vipuniversalvisualreward,
	title        = {{VIP: Towards Universal Visual Reward and Representation via Value-Implicit Pre-Training}},
	author       = {Yecheng Jason Ma and Shagun Sodhani and Dinesh Jayaraman and Osbert Bastani and Vikash Kumar and Amy Zhang},
	year         = 2023,
	url          = {https://arxiv.org/abs/2210.00030},
	eprint       = {2210.00030},
	archiveprefix = {arXiv},
	primaryclass = {cs.RO}, 
    note = {arXiv:2210.00030 [cs.RO]}
}

@inproceedings{wang2024rlvlmf,
	title        = {{RL-VLM-F: Reinforcement Learning from Vision Language Foundation Model Feedback}},
	author       = {Wang, Yufei and Sun, Zhanyi and Zhang, Jesse and Xian, Zhou and Biyik, Erdem and Held, David and Erickson, Zackory},
	year         = 2024,
	booktitle    = {Proceedings of the 41th International Conference on Machine Learning}
}

@inproceedings{oneill2024openx,
	title        = {{Open X-Embodiment: Robotic Learning Datasets and RT-X Models: Open X-Embodiment Collaboration}},
	author       = {O’Neill, Abby and Rehman, Abdul and Maddukuri, Abhiram and Gupta, Abhishek and Padalkar, Abhishek and Lee, Abraham and Pooley, Acorn and Gupta, Agrim and Mandlekar, Ajay and Jain, Ajinkya and Tung, Albert and Bewley, Alex and Herzog, Alex and Irpan, Alex and Khazatsky, Alexander and Rai, Anant and Gupta, Anchit and Wang, Andrew and Singh, Anikait and Garg, Animesh and Kembhavi, Aniruddha and Xie, Annie and Brohan, Anthony and Raffin, Antonin and Sharma, Archit and Yavary, Arefeh and Jain, Arhan and Balakrishna, Ashwin and Wahid, Ayzaan and Burgess-Limerick, Ben and Kim, Beomjoon and Schölkopf, Bernhard and Wulfe, Blake and Ichter, Brian and Lu, Cewu and Xu, Charles and Le, Charlotte and Finn, Chelsea and Wang, Chen and Xu, Chenfeng and Chi, Cheng and Huang, Chenguang and Chan, Christine and Agia, Christopher and Pan, Chuer and Fu, Chuyuan and Devin, Coline and Xu, Danfei and Morton, Daniel and Driess, Danny and Chen, Daphne and Pathak, Deepak and Shah, Dhruv and Büchler, Dieter and Jayaraman, Dinesh and Kalashnikov, Dmitry and Sadigh, Dorsa and Johns, Edward and Foster, Ethan and Liu, Fangchen and Ceola, Federico and Xia, Fei and Zhao, Feiyu and Stulp, Freek and Zhou, Gaoyue and Sukhatme, Gaurav S. and Salhotra, Gautam and Yan, Ge and Feng, Gilbert and Schiavi, Giulio and Berseth, Glen and Kahn, Gregory and Wang, Guanzhi and Su, Hao and Fang, Hao-Shu and Shi, Haochen and Bao, Henghui and Ben Amor, Heni and Christensen, Henrik I and Furuta, Hiroki and Walke, Homer and Fang, Hongjie and Ha, Huy and Mordatch, Igor and Radosavovic, Ilija and Leal, Isabel and Liang, Jacky and Abou-Chakra, Jad and Kim, Jaehyung and Drake, Jaimyn and Peters, Jan and Schneider, Jan and Hsu, Jasmine and Bohg, Jeannette and Bingham, Jeffrey and Wu, Jeffrey and Gao, Jensen and Hu, Jiaheng and Wu, Jiajun and Wu, Jialin and Sun, Jiankai and Luo, Jianlan and Gu, Jiayuan and Tan, Jie and Oh, Jihoon and Wu, Jimmy and Lu, Jingpei and Yang, Jingyun and Malik, Jitendra and Silvério, João and Hejna, Joey and Booher, Jonathan and Tompson, Jonathan and Yang, Jonathan and Salvador, Jordi and Lim, Joseph J. and Han, Junhyek and Wang, Kaiyuan and Rao, Kanishka and Pertsch, Karl and Hausman, Karol and Go, Keegan and Gopalakrishnan, Keerthana and Goldberg, Ken and Byrne, Kendra and Oslund, Kenneth and Kawaharazuka, Kento and Black, Kevin and Lin, Kevin and Zhang, Kevin and Ehsani, Kiana and Lekkala, Kiran and Ellis, Kirsty and Rana, Krishan and Srinivasan, Krishnan and Fang, Kuan and Singh, Kunal Pratap and Zeng, Kuo-Hao and Hatch, Kyle and Hsu, Kyle and Itti, Laurent and Chen, Lawrence Yunliang and Pinto, Lerrel and Fei-Fei, Li and Tan, Liam and Fan, Linxi Jim and Ott, Lionel and Lee, Lisa and Weihs, Luca and Chen, Magnum and Lepert, Marion and Memmel, Marius and Tomizuka, Masayoshi and Itkina, Masha and Castro, Mateo Guaman and Spero, Max and Du, Maximilian and Ahn, Michael and Yip, Michael C. and Zhang, Mingtong and Ding, Mingyu and Heo, Minho and Srirama, Mohan Kumar and Sharma, Mohit and Kim, Moo Jin and Kanazawa, Naoaki and Hansen, Nicklas and Heess, Nicolas and Joshi, Nikhil J and Suenderhauf, Niko and Liu, Ning and Di Palo, Norman and Shafiullah, Nur Muhammad Mahi and Mees, Oier and Kroemer, Oliver and Bastani, Osbert and Sanketi, Pannag R and Miller, Patrick Tree and Yin, Patrick and Wohlhart, Paul and Xu, Peng and Fagan, Peter David and Mitrano, Peter and Sermanet, Pierre and Abbeel, Pieter and Sundaresan, Priya and Chen, Qiuyu and Vuong, Quan and Rafailov, Rafael and Tian, Ran and Doshi, Ria and Martín-Martín, Roberto and Baijal, Rohan and Scalise, Rosario and Hendrix, Rose and Lin, Roy and Qian, Runjia and Zhang, Ruohan and Mendonca, Russell and Shah, Rutav and Hoque, Ryan and Julian, Ryan and Bustamante, Samuel and Kirmani, Sean and Levine, Sergey and Lin, Shan and Moore, Sherry and Bahl, Shikhar and Dass, Shivin and Sonawani, Shubham and Song, Shuran and Xu, Sichun and Haldar, Siddhant and Karamcheti, Siddharth and Adebola, Simeon and Guist, Simon and Nasiriany, Soroush and Schaal, Stefan and Welker, Stefan and Tian, Stephen and Ramamoorthy, Subramanian and Dasari, Sudeep and Belkhale, Suneel and Park, Sungjae and Nair, Suraj and Mirchandani, Suvir and Osa, Takayuki and Gupta, Tanmay and Harada, Tatsuya and Matsushima, Tatsuya and Xiao, Ted and Kollar, Thomas and Yu, Tianhe and Ding, Tianli and Davchev, Todor and Zhao, Tony Z. and Armstrong, Travis and Darrell, Trevor and Chung, Trinity and Jain, Vidhi and Vanhoucke, Vincent and Zhan, Wei and Zhou, Wenxuan and Burgard, Wolfram and Chen, Xi and Wang, Xiaolong and Zhu, Xinghao and Geng, Xinyang and Liu, Xiyuan and Liangwei, Xu and Li, Xuanlin and Lu, Yao and Ma, Yecheng Jason and Kim, Yejin and Chebotar, Yevgen and Zhou, Yifan and Zhu, Yifeng and Wu, Yilin and Xu, Ying and Wang, Yixuan and Bisk, Yonatan and Cho, Yoonyoung and Lee, Youngwoon and Cui, Yuchen and Cao, Yue and Wu, Yueh-Hua and Tang, Yujin and Zhu, Yuke and Zhang, Yunchu and Jiang, Yunfan and Li, Yunshuang and Li, Yunzhu and Iwasawa, Yusuke and Matsuo, Yutaka and Ma, Zehan and Xu, Zhuo and Cui, Zichen Jeff and Zhang, Zichen and Lin, Zipeng},
	year         = 2024,
	booktitle    = {2024 IEEE International Conference on Robotics and Automation (ICRA)},
	volume       = {},
	number       = {},
	pages        = {6892--6903},
	doi          = {10.1109/ICRA57147.2024.10611477}
}

@inproceedings{kim2025openvla,
	title        = {{OpenVLA: An Open-Source Vision-Language-Action Model}},
	author       = {Kim, Moo Jin and Pertsch, Karl and Karamcheti, Siddharth and Xiao, Ted and Balakrishna, Ashwin and Nair, Suraj and Rafailov, Rafael and Foster, Ethan P and Sanketi, Pannag R and Vuong, Quan and Kollar, Thomas and Burchfiel, Benjamin and Tedrake, Russ and Sadigh, Dorsa and Levine, Sergey and Liang, Percy and Finn, Chelsea},
	year         = 2025,
	booktitle    = {Proceedings of The 8th Conference on Robot Learning},
	publisher    = {PMLR},
	series       = {Proceedings of Machine Learning Research},
	volume       = 270,
	pages        = {2679--2713},
	url          = {https://proceedings.mlr.press/v270/kim25c.html},
	editor       = {Agrawal, Pulkit and Kroemer, Oliver and Burgard, Wolfram}
}

@inproceedings{meng2024simpo,
	title        = {Sim{PO}: {Simple Preference Optimization with a Reference-Free Reward}},
	author       = {Yu Meng and Mengzhou Xia and Danqi Chen},
	year         = 2024,
	booktitle    = {The Thirty-eighth Annual Conference on Neural Information Processing Systems},
	url          = {https://openreview.net/forum?id=3Tzcot1LKb}
}

@inproceedings{rafailov2023dpo,
	title        = {{Direct Preference Optimization}: {{Your}} {Language Model Is Secretly a Reward Model}},
	author       = {Rafailov, Rafael and Sharma, Archit and Mitchell, Eric and Manning, Christopher D and Ermon, Stefano and Finn, Chelsea},
	year         = 2023,
	booktitle    = {Advances in Neural Information Processing Systems},
	publisher    = {Curran Associates, Inc.},
	volume       = 36,
	pages        = {53728--53741},
	editor       = {Oh, A. and Naumann, T. and Globerson, A. and Saenko, K. and Hardt, M. and Levine, S.}
}

@inproceedings{christiano2017rlhf,
	title        = {{Deep Reinforcement Learning from Human Preferences}},
	author       = {Christiano, Paul F and Leike, Jan and Brown, Tom and Martic, Miljan and Legg, Shane and Amodei, Dario},
	year         = 2017,
	booktitle    = {Advances in Neural Information Processing Systems},
	publisher    = {Curran Associates, Inc.},
	volume       = 30,
	pages        = {},
	editor       = {I. Guyon and U. Von Luxburg and S. Bengio and H. Wallach and R. Fergus and S. Vishwanathan and R. Garnett}
}

@article{ouyang2022TrainingLanguageModels,
	title        = {Training Language Models to Follow Instructions with Human Feedback},
	author       = {Ouyang, Long and Wu, Jeffrey and Jiang, Xu and Almeida, Diogo and Wainwright, Carroll and Mishkin, Pamela and Zhang, Chong and Agarwal, Sandhini and Slama, Katarina and Ray, Alex and Schulman, John and Hilton, Jacob and Kelton, Fraser and Miller, Luke and Simens, Maddie and Askell, Amanda and Welinder, Peter and Christiano, Paul F. and Leike, Jan and Lowe, Ryan},
	year         = 2022,
	month        = dec,
	journal      = {Advances in Neural Information Processing Systems},
	volume       = 35,
	pages        = {27730--27744},
	urldate      = {2023-09-26},
	langid       = {english}
}

@misc{openaiGPT4TechnicalReport2023,
	title        = {{GPT-4 Technical Report}},
	author       = {OpenAI},
	year         = 2023,
	note         = {arXiv preprint arXiv:2303.08774 [cs.CL]},
	eprint       = {2303.08774},
	archiveprefix = {arXiv},
	primaryclass = {cs.AI}
}

@article{meta2024llama,
	title        = {{The Llama 3 Herd of Models}},
	author       = {Abhimanyu Dubey and Abhinav Jauhri and Abhinav Pandey and Abhishek Kadian and Ahmad Al-Dahle and Aiesha Letman and Akhil Mathur and Alan Schelten and Amy Yang and Angela Fan and Anirudh Goyal and Anthony Hartshorn and Aobo Yang and Archi Mitra and Archie Sravankumar and Artem Korenev and Arthur Hinsvark and Arun Rao and Aston Zhang and Aurélien Rodriguez and Austen Gregerson and Ava Spataru and Baptiste Rozière and Bethany Biron and Binh Tang and Bobbie Chern and Charlotte Caucheteux and Chaya Nayak and Chloe Bi and Chris Marra and Chris McConnell and Christian Keller and Christophe Touret and Chunyang Wu and Corinne Wong and Cristian Canton Ferrer and Cyrus Nikolaidis and Damien Allonsius and Daniel Song and Danielle Pintz and Danny Livshits and David Esiobu and Dhruv Choudhary and Dhruv Mahajan and Diego Garcia-Olano and Diego Perino and Dieuwke Hupkes and Egor Lakomkin and Ehab AlBadawy and Elina Lobanova and Emily Dinan and Eric Michael Smith and Filip Radenovic and Frank Zhang and Gabriel Synnaeve and Gabrielle Lee and Georgia Lewis Anderson and Graeme Nail and Grégoire Mialon and Guan Pang and Guillem Cucurell and Hailey Nguyen and Hannah Korevaar and Hu Xu and Hugo Touvron and Iliyan Zarov and Imanol Arrieta Ibarra and Isabel M. Kloumann and Ishan Misra and Ivan Evtimov and Jade Copet and Jaewon Lee and Jan Geffert and Jana Vranes and Jason Park and Jay Mahadeokar and Jeet Shah and Jelmer van der Linde and Jennifer Billock and Jenny Hong and Jenya Lee and Jeremy Fu and Jianfeng Chi and Jianyu Huang and Jiawen Liu and Jie Wang and Jiecao Yu and Joanna Bitton and Joe Spisak and Jongsoo Park and Joseph Rocca and Joshua Johnstun and Joshua Saxe and Junteng Jia and Kalyan Vasuden Alwala and Kartikeya Upasani and Kate Plawiak and Ke Li and Kenneth Heafield and Kevin Stone and et al.},
	year         = 2024,
	journal      = {CoRR},
	volume       = {abs/2407.21783},
	url          = {https://doi.org/10.48550/arXiv.2407.21783},
	publtype     = {informal},
	cdate        = 1704067200000
}

@misc{wulff2025jointactionlanguagemodelling,
	title        = {{Joint Action Language Modelling for Transparent Policy Execution}},
	author       = {Theodor Wulff and Rahul Singh Maharjan and Xinyun Chi and Angelo Cangelosi},
	year         = 2025,
	url          = {https://arxiv.org/abs/2504.10055},
	eprint       = {2504.10055},
	archiveprefix = {arXiv},
	primaryclass = {cs.RO}, 
    note = {arXiv:2504.10055 [cs.RO]}
}

@InProceedings{radfordLearningTransferableVisual2021,
  title = 	 {{Learning Transferable Visual Models From Natural Language Supervision}},
  author =       {Radford, Alec and Kim, Jong Wook and Hallacy, Chris and Ramesh, Aditya and Goh, Gabriel and Agarwal, Sandhini and Sastry, Girish and Askell, Amanda and Mishkin, Pamela and Clark, Jack and Krueger, Gretchen and Sutskever, Ilya},
  booktitle = 	 {Proceedings of the 38th International Conference on Machine Learning},
  pages = 	 {8748--8763},
  year = 	 {2021},
  editor = 	 {Meila, Marina and Zhang, Tong},
  volume = 	 {139},
  series = 	 {Proceedings of Machine Learning Research},
  month = 	 {18--24 Jul},
  publisher =    {PMLR},
  url = 	 {https://proceedings.mlr.press/v139/radford21a.html}
}

@article{gr00tn1_2025,
  publtype={informal},
  author={Johan Bjorck and Fernando Castañeda and Nikita Cherniadev and Xingye Da and Runyu Ding and Linxi and Yu Fang and Dieter Fox and Fengyuan Hu and Spencer Huang and Joel Jang and Zhenyu Jiang and Jan Kautz and Kaushil Kundalia and Lawrence Lao and Zhiqi Li and Zongyu Lin and Kevin Lin and Guilin Liu and Edith LLontop and Loic Magne and Ajay Mandlekar and Avnish Narayan and Soroush Nasiriany and Scott Reed and You Liang Tan and Guanzhi Wang and Zu Wang and Jing Wang and Qi Wang and Jiannan Xiang and Yuqi Xie and Yinzhen Xu and Zhenjia Xu and Seonghyeon Ye and Zhiding Yu and Ao Zhang and Hao Zhang and Yizhou Zhao and Ruijie Zheng and Yuke Zhu},
  title={{GR00T N1: An Open Foundation Model for Generalist Humanoid Robots}},
  year={2025},
  month={March},
  cdate={1740787200000},
  journal={CoRR},
  volume={abs/2503.14734},
  url={https://doi.org/10.48550/arXiv.2503.14734}
}

@inproceedings{shiHiRobotOpenEnded2025,
title={{Hi Robot: Open-Ended Instruction Following with Hierarchical Vision-Language-Action Models}},
author={Lucy Xiaoyang Shi and brian ichter and Michael Robert Equi and Liyiming Ke and Karl Pertsch and Quan Vuong and James Tanner and Anna Walling and Haohuan Wang and Niccolo Fusai and Adrian Li-Bell and Danny Driess and Lachy Groom and Sergey Levine and Chelsea Finn},
booktitle={Forty-second International Conference on Machine Learning (ICML)},
year={2025},
url={https://openreview.net/forum?id=lNVHg9npif}
}

@inproceedings{belkhaleRTHActionHierarchies2024,
  author={Suneel Belkhale and Tianli Ding and Ted Xiao and Pierre Sermanet and Quan Vuong and Jonathan Tompson and Yevgen Chebotar and Debidatta Dwibedi and Dorsa Sadigh},
  title={{RT-H: Action Hierarchies using Language}},
  year={2024},
  cdate={1704067200000},
  url={https://doi.org/10.15607/RSS.2024.XX.049},
  booktitle={Robotics: Science and Systems},
}

@inproceedings{blackpi05VisionLanguageActionModel2025,
	title        = {$\pi_{0.5}$: A {{Vision-Language-Action Model}} with {{Open-World Generalization}}},
	author       = {Black, Kevin and Brown, Noah and Darpinian, James and Dhabalia, Karan and Driess, Danny and Esmail, Adnan and Equi, Michael Robert and Finn, Chelsea and Fusai, Niccolo and Galliker, Manuel Y. and Ghosh, Dibya and Groom, Lachy and Hausman, Karol and Ichter, Brian and Jakubczak, Szymon and Jones, Tim and Ke, Liyiming and LeBlanc, Devin and Levine, Sergey and {Li-Bell}, Adrian and Mothukuri, Mohith and Nair, Suraj and Pertsch, Karl and Ren, Allen Z. and Shi, Lucy Xiaoyang and Smith, Laura and Springenberg, Jost Tobias and Stachowicz, Kyle and Tanner, James and Vuong, Quan and Walke, Homer and Walling, Anna and Wang, Haohuan and Yu, Lili and Zhilinsky, Ury},
	year         = 2025,
	month        = sep,
	booktitle    = {9th {{Annual Conference}} on {{Robot Learning}} (CoRL)},
	urldate      = {2025-10-23},
	langid       = {english}
}

@article{teamGemma3Technical2025,
  publtype={informal},
  author={Aishwarya Kamath and Johan Ferret and Shreya Pathak and Nino Vieillard and Ramona Merhej and Sarah Perrin and Tatiana Matejovicova and Alexandre Ramé and Morgane Rivière and Louis Rouillard and Thomas Mesnard and Geoffrey Cideron and Jean-Bastien Grill and Sabela Ramos and Edouard Yvinec and Michelle Casbon and Etienne Pot and Ivo Penchev and Gaël Liu and Francesco Visin and Kathleen Kenealy and Lucas Beyer and Xiaohai Zhai and Anton Tsitsulin and Róbert Busa-Fekete and Alex Feng and Noveen Sachdeva and Benjamin Coleman and Yi Gao and Basil Mustafa and Iain Barr and Emilio Parisotto and David Tian and Matan Eyal and Colin Cherry and Jan-Thorsten Peter and Danila Sinopalnikov and Surya Bhupatiraju and Rishabh Agarwal and Mehran Kazemi and Dan Malkin and Ravin Kumar and David Vilar and Idan Brusilovsky and Jiaming Luo and Andreas Steiner and Abe Friesen and Abhanshu Sharma and Abheesht Sharma and Adi Mayrav Gilady and Adrian Goedeckemeyer and Alaa Saade and Alexander Kolesnikov and Alexei Bendebury and Alvin Abdagic and Amit Vadi and András György and André Susano Pinto and Anil Das and Ankur Bapna and Antoine Miech and Antoine Yang and Antonia Paterson and Ashish Shenoy and Ayan Chakrabarti and Bilal Piot and Bo Wu and Bobak Shahriari and Bryce Petrini and Charlie Chen and Charline Le Lan and Christopher A. Choquette-Choo and CJ Carey and Cormac Brick and Daniel Deutsch and Danielle Eisenbud and Dee Cattle and Derek Cheng and Dimitris Paparas and Divyashree Shivakumar Sreepathihalli and Doug Reid and Dustin Tran and Dustin Zelle and Eric Noland and Erwin Huizenga and Eugene Kharitonov and Frederick Liu and Gagik Amirkhanyan and Glenn Cameron and Hadi Hashemi and Hanna Klimczak-Plucinska and Harman Singh and Harsh Mehta and Harshal Tushar Lehri and Hussein Hazimeh and Ian Ballantyne and Idan Szpektor and Ivan Nardini},
  title={{Gemma 3 Technical Report}},
  year={2025},
  month={March},
  cdate={1740787200000},
  journal={CoRR},
  volume={abs/2503.19786},
  url={https://doi.org/10.48550/arXiv.2503.19786}
}

@ARTICLE{lynchInteractiveLanguageTalking2022,
  author={Lynch, Corey and Wahid, Ayzaan and Tompson, Jonathan and Ding, Tianli and Betker, James and Baruch, Robert and Armstrong, Travis and Florence, Pete},
  journal={IEEE Robotics and Automation Letters}, 
  title={{Interactive Language: Talking to Robots in Real Time}}, 
  year={2023},
  volume={},
  number={},
  pages={1-8},
  doi={10.1109/LRA.2023.3295255}
}

@misc{shukorSmolVLAVisionLanguageActionModel2025a,
	title        = {{{SmolVLA}}: {{A Vision-Language-Action Model}} for {{Affordable}} and {{Efficient Robotics}}},
	shorttitle   = {{SmolVLA}},
	author       = {Shukor, Mustafa and Aubakirova, Dana and Capuano, Francesco and Kooijmans, Pepijn and Palma, Steven and Zouitine, Adil and Aractingi, Michel and Pascal, Caroline and Russi, Martino and Marafioti, Andres and Alibert, Simon and Cord, Matthieu and Wolf, Thomas and Cadene, Remi},
	year         = 2025,
	month        = jun,
	publisher    = {arXiv},
	number       = {arXiv:2506.01844},
	doi          = {10.48550/arXiv.2506.01844},
	urldate      = {2025-08-12},
	eprint       = {2506.01844},
	primaryclass = {cs},
	archiveprefix = {arXiv}, 
    note = {arXiv:2506.01844 [cs.RO]}
}

@article{black2024p0VisionLanguageActionFlow,
  publtype={informal},
  author={Kevin Black and Noah Brown and Danny Driess and Adnan Esmail and Michael Equi and Chelsea Finn and Niccolo Fusai and Lachy Groom and Karol Hausman and Brian Ichter and Szymon Jakubczak and Tim Jones and Liyiming Ke and Sergey Levine and Adrian Li-Bell and Mohith Mothukuri and Suraj Nair and Karl Pertsch and Lucy Xiaoyang Shi and James Tanner and Quan Vuong and Anna Walling and Haohuan Wang and Ury Zhilinsky},
  title={$\pi_0$: {A Vision-Language-Action Flow Model for General Robot Control}},
  year={2024},
  cdate={1704067200000},
  journal={CoRR},
  volume={abs/2410.24164},
  url={https://doi.org/10.48550/arXiv.2410.24164}
}

@article{awaisFoundationModelsDefining2025,
	title        = {Foundation {{Models Defining}} a {{New Era}} in {{Vision}}: {{A Survey}} and {{Outlook}}},
	shorttitle   = {Foundation {{Models Defining}} a {{New Era}} in {{Vision}}},
	author       = {Awais, Muhammad and Naseer, Muzammal and Khan, Salman and Anwer, Rao Muhammad and Cholakkal, Hisham and Shah, Mubarak and Yang, Ming-Hsuan and Khan, Fahad Shahbaz},
	year         = 2025,
	month        = apr,
	journal      = {IEEE Transactions on Pattern Analysis and Machine Intelligence},
	volume       = 47,
	number       = 4,
	pages        = {2245--2264},
	doi          = {10.1109/TPAMI.2024.3506283},
	issn         = {1939-3539},
	urldate      = {2025-11-04}
}

@misc{brohanRT1RoboticsTransformer2023b,
	title        = {{{RT-1}}: {{Robotics Transformer}} for {{Real-World Control}} at {{Scale}}},
	shorttitle   = {{RT-1}},
	author       = {Brohan, Anthony and Brown, Noah and Carbajal, Justice and Chebotar, Yevgen and Dabis, Joseph and Finn, Chelsea and Gopalakrishnan, Keerthana and Hausman, Karol and Herzog, Alex and Hsu, Jasmine and Ibarz, Julian and Ichter, Brian and Irpan, Alex and Jackson, Tomas and Jesmonth, Sally and Joshi, Nikhil J. and Julian, Ryan and Kalashnikov, Dmitry and Kuang, Yuheng and Leal, Isabel and Lee, Kuang-Huei and Levine, Sergey and Lu, Yao and Malla, Utsav and Manjunath, Deeksha and Mordatch, Igor and Nachum, Ofir and Parada, Carolina and Peralta, Jodilyn and Perez, Emily and Pertsch, Karl and Quiambao, Jornell and Rao, Kanishka and Ryoo, Michael and Salazar, Grecia and Sanketi, Pannag and Sayed, Kevin and Singh, Jaspiar and Sontakke, Sumedh and Stone, Austin and Tan, Clayton and Tran, Huong and Vanhoucke, Vincent and Vega, Steve and Vuong, Quan and Xia, Fei and Xiao, Ted and Xu, Peng and Xu, Sichun and Yu, Tianhe and Zitkovich, Brianna},
	year         = 2023,
	month        = aug,
	publisher    = {arXiv},
	number       = {arXiv:2212.06817},
	urldate      = {2023-10-16},
	eprint       = {2212.06817},
	archiveprefix = {arXiv}, 
      note = {arXiv:2212.06817 [cs.RO]}
}

@inproceedings{brohanRT2VisionLanguageActionModels2023a,
    title={{RT}-2: {Vision-Language-Action Models Transfer Web Knowledge to Robotic Control}},
    author={Brianna Zitkovich and Tianhe Yu and Sichun Xu and Peng Xu and Ted Xiao and Fei Xia and Jialin Wu and Paul Wohlhart and Stefan Welker and Ayzaan Wahid and Quan Vuong and Vincent Vanhoucke and Huong Tran and Radu Soricut and Anikait Singh and Jaspiar Singh and Pierre Sermanet and Pannag R Sanketi and Grecia Salazar and Michael S Ryoo and Krista Reymann and Kanishka Rao and Karl Pertsch and Igor Mordatch and Henryk Michalewski and Yao Lu and Sergey Levine and Lisa Lee and Tsang-Wei Edward Lee and Isabel Leal and Yuheng Kuang and Dmitry Kalashnikov and Ryan Julian and Nikhil J Joshi and Alex Irpan and brian ichter and Jasmine Hsu and Alexander Herzog and Karol Hausman and Keerthana Gopalakrishnan and Chuyuan Fu and Pete Florence and Chelsea Finn and Kumar Avinava Dubey and Danny Driess and Tianli Ding and Krzysztof Marcin Choromanski and Xi Chen and Yevgen Chebotar and Justice Carbajal and Noah Brown and Anthony Brohan and Montserrat Gonzalez Arenas and Kehang Han},
    booktitle={7th Annual Conference on Robot Learning (CoRL)},
    year={2023},
    url={https://openreview.net/forum?id=XMQgwiJ7KSX}
}

@inproceedings{octomodelteamOctoOpenSourceGeneralist2024,
    title={{Octo: An Open-Source Generalist Robot Policy}},
    author={Oier Mees and Dibya Ghosh and Karl Pertsch and Kevin Black and Homer Rich Walke and Sudeep Dasari and Joey Hejna and Tobias Kreiman and Charles Xu and Jianlan Luo and You Liang Tan and Dorsa Sadigh and Chelsea Finn and Sergey Levine},
    booktitle={First Workshop on Vision-Language Models for Navigation and Manipulation at ICRA 2024},
    year={2024},
    url={https://openreview.net/forum?id=jGrtIvJBpS}
}

@inproceedings{stepputtisLanguageConditionedImitationLearning2020a,
	title        = {Language-{{Conditioned Imitation Learning}} for {{Robot Manipulation Tasks}}},
	author       = {Stepputtis, Simon and Campbell, Joseph and Phielipp, Mariano and Lee, Stefan and Baral, Chitta and Ben Amor, Heni},
	year         = 2020,
	booktitle    = {Advances in {{Neural Information Processing Systems}}},
	publisher    = {Curran Associates, Inc.},
	volume       = 33,
	pages        = {13139--13150},
	urldate      = {2023-10-09}
}

@inproceedings{nairLearningLanguageConditionedRobot2022a,
	title        = {Learning {{Language-Conditioned Robot Behavior}} from {{Offline Data}} and {{Crowd-Sourced Annotation}}},
	author       = {Nair, Suraj and Mitchell, Eric and Chen, Kevin and Ichter, Brian and Savarese, Silvio and Finn, Chelsea},
	year         = 2022,
	month        = jan,
	booktitle    = {Proceedings of the 5th {{Conference}} on {{Robot Learning}}},
	publisher    = {PMLR},
	pages        = {1303--1315},
	issn         = {2640-3498},
	urldate      = {2024-07-04},
	langid       = {english}
}

@misc{tschannenSigLIP2Multilingual2025,
	title        = {{{SigLIP}} 2: {{Multilingual Vision-Language Encoders}} with {{Improved Semantic Understanding}}, {{Localization}}, and {{Dense Features}}},
	shorttitle   = {{{SigLIP}} 2},
	author       = {Tschannen, Michael and Gritsenko, Alexey and Wang, Xiao and Naeem, Muhammad Ferjad and Alabdulmohsin, Ibrahim and Parthasarathy, Nikhil and Evans, Talfan and Beyer, Lucas and Xia, Ye and Mustafa, Basil and H{\'e}naff, Olivier and Harmsen, Jeremiah and Steiner, Andreas and Zhai, Xiaohua},
	year         = 2025,
	month        = feb,
	publisher    = {arXiv},
	number       = {arXiv:2502.14786},
	doi          = {10.48550/arXiv.2502.14786},
	urldate      = {2025-07-24},
	eprint       = {2502.14786},
	primaryclass = {cs},
	archiveprefix = {arXiv},
    note = {arXiv:2502.14786 [cs.CV]}
}

@article{perezFiLMVisualReasoning2018,
	title        = {{{FiLM}}: {{Visual Reasoning}} with a {{General Conditioning Layer}}},
	author       = {Perez, Ethan and Strub, Florian and {de Vries}, Harm and Dumoulin, Vincent and Courville, Aaron},
	year         = 2018,
	month        = apr,
	journal      = {Proceedings of the AAAI Conference on Artificial Intelligence},
	volume       = 32,
	number       = 1,
	doi          = {10.1609/aaai.v32i1.11671}
}

@article{heMomentumContrastUnsupervised,
	title        = {Momentum {{Contrast}} for {{Unsupervised Visual Representation Learning}}},
	author       = {He, Kaiming and Fan, Haoqi and Wu, Yuxin and Xie, Saining and Girshick, Ross},
	year         = 2020,
	journal      = {CVPR},
	langid       = {english}
}

@inproceedings{ma2025contrastive,
	title        = {{Contrastive Imitation Learning for Language-guided Multi-Task Robotic Manipulation}},
	author       = {Ma, Teli and Zhou, Jiaming and Wang, Zifan and Qiu, Ronghe and Liang, Junwei},
	year         = 2025,
	month        = {06--09 Nov},
	booktitle    = {Proceedings of The 8th Conference on Robot Learning},
	publisher    = {PMLR},
	series       = {Proceedings of Machine Learning Research},
	volume       = 270,
	pages        = {4651--4669},
	url          = {https://proceedings.mlr.press/v270/ma25a.html},
	editor       = {Agrawal, Pulkit and Kroemer, Oliver and Burgard, Wolfram}
}

@inproceedings{zhaiSigmoidLossLanguage2023,
	title        = {Sigmoid {{Loss}} for {{Language Image Pre-Training}}},
	author       = {Zhai, Xiaohua and Mustafa, Basil and Kolesnikov, Alexander and Beyer, Lucas},
	year         = 2023,
	month        = oct,
	booktitle    = {2023 {{IEEE}}/{{CVF International Conference}} on {{Computer Vision}} ({{ICCV}})},
	publisher    = {IEEE},
	address      = {Paris, France},
	pages        = {11941--11952},
	doi          = {10.1109/ICCV51070.2023.01100},
	isbn         = {979-8-3503-0718-4},
	urldate      = {2024-07-12},
	copyright    = {https://doi.org/10.15223/policy-029},
	langid       = {english}
}

@inproceedings{chenSimpleFrameworkContrastive2020,
	title        = {A {{Simple Framework}} for {{Contrastive Learning}} of {{Visual Representations}}},
	author       = {Chen, Ting and Kornblith, Simon and Norouzi, Mohammad and Hinton, Geoffrey},
	year         = 2020,
	month        = nov,
	booktitle    = {Proceedings of the 37th {{International Conference}} on {{Machine Learning}}},
	publisher    = {PMLR},
	pages        = {1597--1607},
	issn         = {2640-3498},
	urldate      = {2024-03-04},
	langid       = {english}
}

@misc{ma2024survey,
      title={{A Survey on Vision-Language-Action Models for Embodied AI}}, 
      author={Yueen Ma and Zixing Song and Yuzheng Zhuang and Jianye Hao and Irwin King},
      year={2025},
      eprint={2405.14093},
      archivePrefix={arXiv},
      primaryClass={cs.RO},
      url={https://arxiv.org/abs/2405.14093},
      note = {arXiv:2405.14093 [cs.RO]}
}

@inproceedings{eysenbachContrastiveLearningGoalConditioned2022,
	title        = {{Contrastive Learning as Goal-Conditioned Reinforcement Learning}},
	author       = {Eysenbach, Benjamin and Zhang, Tianjun and Levine, Sergey and Salakhutdinov, Ruslan},
	year         = 2022,
	booktitle    = {{Advances in Neural Information Processing Systems 35 - 36th Conference on Neural Information Processing Systems, NeurIPS 2022}},
	publisher    = {Neural information processing systems foundation},
	urldate      = {2025-11-06},
	langid       = {English (US)}
}

@inproceedings{yangRank2RewardLearningShaped2024,
	title        = {{{Rank2Reward}}: {{Learning Shaped Reward Functions}} from {{Passive Video}}},
	shorttitle   = {{Rank2Reward}},
	author       = {Yang, Daniel and Tjia, Davin and Berg, Jacob and Damen, Dima and Agrawal, Pulkit and Gupta, Abhishek},
	year         = 2024,
	month        = may,
	booktitle    = {2024 {{IEEE International Conference}} on {{Robotics}} and {{Automation}} ({{ICRA}})},
	pages        = {2806--2813},
	doi          = {10.1109/ICRA57147.2024.10610873},
	urldate      = {2024-10-29}
}

@inproceedings{bizaOnRobotReinforcementLearning2025,
	title        = {On-{{Robot Reinforcement Learning}} with {{Goal-Contrastive Rewards}}},
	author       = {Biza, Ondrej and Weng, Thomas and Sun, Lingfeng and Schmeckpeper, Karl and Kelestemur, Tarik and Ma, Yecheng Jason and Platt, Robert and {van de Meent}, Jan-Willem and Wong, Lawson L.S.},
	year         = 2025,
	month        = may,
	booktitle    = {2025 {{IEEE International Conference}} on {{Robotics}} and {{Automation}} ({{ICRA}})},
	pages        = {4797--4805},
	doi          = {10.1109/ICRA55743.2025.11128466},
	urldate      = {2025-11-06}
}

@inproceedings{
palo2024keypoint,
title={{Keypoint Action Tokens Enable In-Context Imitation Learning in Robotics}},
author={Norman Di Palo and Edward Johns},
booktitle={First Workshop on Vision-Language Models for Navigation and Manipulation at ICRA 2024},
year={2024},
url={https://openreview.net/forum?id=6QKttpbRS1}
}

@ARTICLE{salehzade2022purposefulComm,
  author={Salehzadeh, Roya and Gong, Jiaqi and Jalili, Nader},
  journal={IEEE Access}, 
  title={{Purposeful Communication in Human–Robot Collaboration: A Review of Modern Approaches in Manufacturing}}, 
  year={2022},
  volume={10},
  number={},
  pages={129344-129361},
  doi={10.1109/ACCESS.2022.3227049}
}

@misc{balestriero2023cookbookselfsupervisedlearning,
      title={{A Cookbook of Self-Supervised Learning}}, 
      author={Randall Balestriero and Mark Ibrahim and Vlad Sobal and Ari Morcos and Shashank Shekhar and Tom Goldstein and Florian Bordes and Adrien Bardes and Gregoire Mialon and Yuandong Tian and Avi Schwarzschild and Andrew Gordon Wilson and Jonas Geiping and Quentin Garrido and Pierre Fernandez and Amir Bar and Hamed Pirsiavash and Yann LeCun and Micah Goldblum},
      year={2023},
      eprint={2304.12210},
      archivePrefix={arXiv},
      primaryClass={cs.LG},
      url={https://arxiv.org/abs/2304.12210}, 
      note = {arXiv:2304.12210 [cs.LG]}
}

@inproceedings{cheng2024spatialrgpt,
 author = {Cheng, An-Chieh and Yin, Hongxu and Fu, Yang and Guo, Qiushan and Yang, Ruihan and Kautz, Jan and Wang, Xiaolong and Liu, Sifei},
 booktitle = {Advances in Neural Information Processing Systems},
 doi = {10.52202/079017-4293},
 editor = {A. Globerson and L. Mackey and D. Belgrave and A. Fan and U. Paquet and J. Tomczak and C. Zhang},
 pages = {135062--135093},
 publisher = {Curran Associates, Inc.},
 title = {{SpatialRGPT: Grounded Spatial Reasoning in Vision-Language Models}},
 url = {https://proceedings.neurips.cc/paper_files/paper/2024/file/f38cb4cf9a5eaa92b3cfa481832719c6-Paper-Conference.pdf},
 volume = {37},
 year = {2024}
}

@inproceedings{lin-2004-rouge,
    title = "{{ROUGE: A Package for Automatic Evaluation of Summaries}}",
    author = "Lin, Chin-Yew",
    booktitle = "Text Summarization Branches Out",
    month = jul,
    year = "2004",
    address = "Barcelona, Spain",
    publisher = "Association for Computational Linguistics",
    url = "https://aclanthology.org/W04-1013/",
    pages = "74--81"
}

@inproceedings{banerjee2005meteor,
  title={{METEOR: An automatic metric for MT evaluation with improved correlation with human judgments}},
  author={Banerjee, Satanjeev and Lavie, Alon},
  booktitle={Proceedings of the acl workshop on intrinsic and extrinsic evaluation measures for machine translation and/or summarization},
  year={2005}
}

@inproceedings{papipeni2002bleu,
author = {Papineni, Kishore and others},
title = {{BLEU: a method for automatic evaluation of machine translation}},
year = {2002},
publisher = {Association for Computational Linguistics},
booktitle = {Proceedings of the 40th Annual Meeting on Association for Computational Linguistics},
}

@inproceedings{zhang2020BERTScore,
title={{BERTScore: Evaluating Text Generation with BERT}},
author={Tianyi Zhang* and Varsha Kishore* and Felix Wu* and Kilian Q. Weinberger and Yoav Artzi},
booktitle={International Conference on Learning Representations},
year={2020},
url={https://openreview.net/forum?id=SkeHuCVFDr}
}

@inproceedings{NEURIPS2023_ed3fea90,
  title = {{Language Models Don't Always Say What They Think: {{Unfaithful}} Explanations in Chain-of-Thought Prompting}},
  booktitle = {Advances in Neural Information Processing Systems},
  author = {Turpin, Miles and Michael, Julian and Perez, Ethan and Bowman, Samuel},
  editor = {Oh, A. and Naumann, T. and Globerson, A. and Saenko, K. and Hardt, M. and Levine, S.},
  year = 2023,
  volume = {36},
  pages = {74952--74965},
  publisher = {Curran Associates, Inc.}
}

@inproceedings{
zhou2023leasttomost,
title={{Least-to-Most Prompting Enables Complex Reasoning in Large Language Models}},
author={Denny Zhou and Nathanael Sch{\"a}rli and Le Hou and Jason Wei and Nathan Scales and Xuezhi Wang and Dale Schuurmans and Claire Cui and Olivier Bousquet and Quoc V Le and Ed H. Chi},
booktitle={The Eleventh International Conference on Learning Representations },
year={2023},
url={https://openreview.net/forum?id=WZH7099tgfM}
}

@article{tervenComprehensiveSurveyLoss2025,
  title = {{A Comprehensive Survey of Loss Functions and Metrics in Deep Learning}},
  author = {Terven, Juan and {Cordova-Esparza}, Diana-Margarita and {Romero-Gonz{\'a}lez}, Julio-Alejandro and {Ram{\'i}rez-Pedraza}, Alfonso and {Ch{\'a}vez-Urbiola}, E. A.},
  year = 2025,
  month = apr,
  journal = {Artificial Intelligence Review},
  volume = {58},
  number = {7},
  pages = {195},
  issn = {1573-7462},
  doi = {10.1007/s10462-025-11198-7},
  urldate = {2025-11-13},
  langid = {english}
}
